\documentclass[journal]{IEEEtran}

\usepackage{cite}
\usepackage{placeins}
\usepackage{tabu}
\usepackage{multirow}
\usepackage[normalem]{ulem} % for striketrough
\usepackage{bm} % bold symbols
\usepackage{color}

\def\zapcolorreset{\let\reset@color\relax\ignorespaces}
\def\colorrows#1{\noalign{\aftergroup\zapcolorreset#1}\ignorespaces}

\ifCLASSINFOpdf
  \usepackage[pdftex]{graphicx}
\else
\fi

\usepackage{amsmath}
\ifCLASSOPTIONcompsoc
  \usepackage[caption=false,font=normalsize,labelfont=sf,textfont=sf]{subfig}
\else
  \usepackage[caption=false,font=footnotesize]{subfig}
\fi

\usepackage{stfloats}

\usepackage{tikz}
\usepackage{mathtools}

\newcommand{\rom}[1]{\uppercase\expandafter{\romannumeral #1\relax}}

\begin{document}

%\begin{center}
%	© 2019 IEEE. Personal use of this material is permitted. Permission from IEEE must be obtained for all other uses, including reprinting/republishing this material for advertising or promotional purposes, collecting new collected works for resale or redistribution to servers or lists, or reuse of any copyrighted component of this work in other works
%\end{center}

\title{Direct Automatic Coronary Calcium Scoring in Cardiac and Chest CT}

\author{Bob D. de Vos, Jelmer M. Wolterink, Tim Leiner, Pim A. de Jong, Nikolas Lessmann, Ivana I\v{s}gum%
	
\thanks{
Copyright © 2019 IEEE. Personal use of this material is permitted. However, permission to use this material for any other purposes must be obtained from the IEEE by sending a request to pubs-permissions@ieee.org.
Bob D. de Vos, Jelmer M. Wolterink, Nikolas Lessmann, and Ivana I\v{s}gum are with the Image Sciences Institute of the University Medical Center Utrecht and Utrecht University, Utrecht, The Netherlands. Tim Leiner and Pim A. de Jong are with the Department of Radiology, University Medical Center Utrecht and Utrecht University, Utrecht, the Netherlands.
This work is part of the research programme ImaGene with project number 12726, which is partly financed by the Netherlands Organisation for Scientific Research (NWO).
The authors thank the National Cancer Institute for access to NCI's data collected by the National Lung Screening Trial. The statements contained herein are solely those of the authors and do not represent or imply concurrence or endorsement by NCI.}% <-this % stops a space
}

\maketitle

\begin{abstract}
Cardiovascular disease (CVD) is the global leading cause of death. A strong risk factor for CVD events is the amount of coronary artery calcium (CAC). To meet demands of the increasing interest in quantification of CAC, i.e. coronary calcium scoring, especially as an unrequested finding for screening and research, automatic methods have been proposed. Current automatic calcium scoring methods are relatively computationally expensive and only provide scores for one type of CT. To address this, we propose a computationally efficient method that employs two ConvNets: the first performs registration to align the fields of view of input CTs and the second performs direct regression of the calcium score, thereby circumventing time-consuming intermediate CAC segmentation. Optional decision feedback provides insight in the regions that contributed to the calcium score. 
Experiments were performed using 903 cardiac CT and 1,687 chest CT scans. The method predicted calcium scores in less than 0.3\,s. Intra-class correlation coefficient between predicted and manual calcium scores was 0.98 for both cardiac and chest CT. 
The method showed almost perfect agreement between automatic and manual CVD risk categorization in both datasets, with a linearly weighted Cohen's kappa of 0.95 in cardiac CT and 0.93 in chest CT. 
Performance is similar to that of state-of-the-art methods, but the proposed method is hundreds of times faster. 
By providing visual feedback, insight is given in the decision process, making it readily implementable in clinical and research settings.
\end{abstract}

\IEEEpeerreviewmaketitle

\section{Introduction}
Cardiovascular disease (CVD) is the global leading cause of death\cite{gbd2016}. To reduce the burden of cardiovascular disease the World Health Organization underlines the need for early detection and treatment of individuals with CVD or those who are at high cardiovascular risk due to the presence of one or more risk factors~\cite{whofactsheet}. A strong and independent risk factor for CVD events, e.g. myocardial infarction, is the quantity of coronary artery calcium (CAC)~\cite{yeboah2012,hecht2015, hecht2017}. Quantification of CAC, i.e. calcium scoring, is typically performed in dedicated non-contrast-enhanced ECG-synchronized cardiac CT scans\cite{hecht2015}. Alternatively, calcium scoring can be performed in other non-contrast-enhanced CTs that visualize the heart; e.g. in low-dose CT attenuation correction scans acquired in hybrid PET/CT and SPECT/CT~\cite{einstein2010,mylonas2012}, or in radiation therapy planning CTs of breast cancer patients~\cite{gernaat2016}. Furthermore, it has been shown that calcium scoring in lung screening low-dose chest CT scans is a predictor for all-cause mortality~\cite{jacobs2010,chiles2015}. In fact, in the National Lung Screening Trial (NLST) CVD was the leading cause of mortality~\cite{nlst2011b}. Thus, CAC quantification, especially as an unrequested finding, has garnered much attention.

Clinically, calcium scoring is performed by experts who manually identify CAC in CT image slices. This is a tedious process of finding and selecting high density voxels in the coronary arteries---commonly defined as two or more connected voxels above 130\,Hounsfield Units (HU). In scans not dedicated to calcium scoring this can be particularly cumbersome because of high noise, low resolution, and motion artifacts. Subsequently, when lesions are identified, region growing is used to fully segment the calcified lesions. Finally, after all CAC lesions have been segmented, CAC is quantified using the Agatston score~\cite{agatston1990}. The Agatston score takes into account the lesion area and the weighted maximum density of the lesion. 
This score can be used to stratify patients into risk categories~\cite{rumberger1999}.

The additional cost involved with manual calcium scoring makes the process prohibitive in settings where it is not the primary request.
To simplify the task, qualitative stratification into CVD risk groups was proposed~\cite{shemesh2010,chiles2015}. Qualitative calcium scoring is faster and it demonstrates good inter-rater agreement. However, such an analysis still demands experts who closely inspect the scans. With the ever-increasing amount of scans and the increasing interest in calcium scoring, especially as an unrequested finding, the use of fully-automatic methods might be the preferred direction. 

Several automatic methods have been introduced for calcium scoring in non-contrast-enhanced CT, ranging from rule-based approaches~\cite{gonzalez2016, xie2017}, to the better performing conventional machine learning approaches\cite{isgum2012,shahzad2013,wolterink2015,durlak2017} and recent deep learning approaches~\cite{wolterink2015miccai,wolterink2016,lessmann2016,lessmann2018}. The main difficulty in automatic calcium scoring is to differentiate CAC from other dense structures. Obviously, CAC exclusively resides in the walls of the coronary arteries, thus most of the automatic methods exploit this prior knowledge.

I\v{s}gum et al.~\cite{isgum2012} introduced the first method for automatic calcium scoring in chest CT. CAC lesions were described with features and subsequently classified using a two-stage classification approach of k-nearest neighbor and support vector classification. Among texture, size, and shape features, highly important for CAC identification, were the location features. Location features were determined by registering an input image to an atlas image and by extracting the location features from a map of a priori spatial probabilities of CAC. The probability map was created from known CAC locations in 237 chest CTs that were registered to a single priorly chosen atlas image. 
Shahzad et al.~\cite{shahzad2013} used a similar machine learning approach for calcium scoring in cardiac CT, but they employed pair-wise deformable image registration to ten atlases that encoded the coronary arteries. The atlases were made from 85 contrast enhanced CT angiography scans with annotated coronary arteries. The methods of I\v{s}gum et al.~\cite{isgum2012} and Shahzad et al.~\cite{shahzad2013} relied on feature selection methods to reduce dimensionality. Wolterink et al.~\cite{wolterink2015} circumvented feature selection by using an extremely randomized trees classifier. Their method also depended on location features that were obtained by deformable image registration of ten atlases with encoded coronary arteries, but these were obtanied from non-contrast-enhanced CTs. 
Durlak et al.~\cite{durlak2017} combined the principles of the aforedescribed methods: they employed a random forest and made an a priori probability map of coronary arteries locations, made from automatically extracted coronary arteries from cardiac CT angiography images. Instead of using time-consuming deformable image registration to align input images and atlas images, they achieved a speed-up by using affine registration. Similarly, other methods employed information from CTA to aid calcium scoring in cardiac CT. These methods were specifically designed for the coronary calcium score (orCaScore) challenge, and employed rule-based image analysis or conventional machine learning \cite{wolterink2016orcascore}.

Most recently proposed methods employ deep learning methods for automatic calcium scoring, in particular convolutional neural networks (ConvNets). ConvNets are known for their automatic feature extracting capabilities and alleviate the need for handcrafting features. Wolterink et al.~\cite{wolterink2016} used ConvNets to classify CAC in cardiac CT angiography scans. All voxels were classified using a pair of ConvNets. One ConvNet identified voxels likely to be CAC and discarded the majority of non-CAC-like voxels such as lung and fatty tissue. The other ConvNet more precisely discriminated between CAC and CAC-like negatives. In the method of Lessmann et al. ~\cite{lessmann2016} a single ConvNet was used that classified candidate CAC lesions in lung screening chest CTs. To simplify the classification tasks, both these deep learning methods used an additional ConvNet that localized the heart with a bounding box~\cite{devos2017localization}. More recently, the method of Lessmann et al.\cite{lessmann2018} fully exploited the feature extraction capabilities of ConvNets without dedicated localization methods. They employed two sequential ConvNets to classify CAC as well as aortic valve, mitral valve, and aorta calcifications in chest CT. The first ConvNet identified candidate calcifications based on their location, and the second ConvNet refined the classification results by reducing false positive errors.

\begin{figure}
    \centering
    \includegraphics[width=\linewidth]{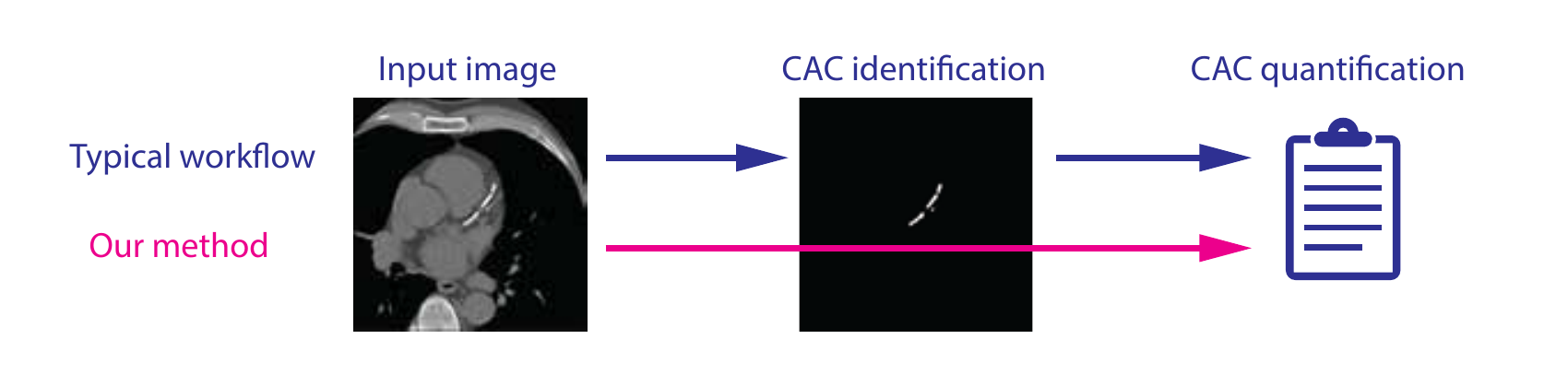}
    \caption{In a typical automatic calcium scoring workflow, CAC is first identified and subsequently quantified. The proposed method uses ConvNet regression to quantify CAC in image slices directly.}
    \label{fig:workflows}
\end{figure}

While all aforementioned methods use different strategies, they all follow a workflow similar to current clinical calcium scoring: CAC is first identified and thereafter quantified. The automatic methods show high accuracy, but often at considerable computational cost. Employing these methods on large datasets would require dedicated servers. To alleviate computational cost, we propose a workflow that circumvents intermediate identification and that performs direct quantification (see Figure~\ref{fig:workflows}). 
Direct quantification has proven to be useful for atrial and ventricle volume quantification~\cite{hussain2017,zhen2017,xue2018}. Furthermore, attempts are being made to use it for calcium scoring. In our preliminary study we presented a direct calcium scoring method that uses 2-D ConvNet regression~\cite{devos2017rsna,devos2017arxiv}. The method performs direct calcium scoring in extracted image slices from bounding boxes cropped around the heart. In a recently proposed method, Cano-Espinosa et al. used a 3-D regression ConvNet for direct calcium scoring in downsampled CT volumes also cropped around the heart. However, their method could not be used in 14\% of the scans, because heart localization failed. Furthermore, previously proposed automatic calcium scoring methods are dedicated to either cardiac CT or chest CT. These methods required retraining for application in other types of CT~\cite{gernaat2016,isgum2017}. 

We present an automatic method that performs real-time direct calcium scoring in different types of non-contrast-enhanced CT. Unlike previous methods that focused on a single type of CT, the proposed method is able to perform calcium scoring directly in multiple types of CT by using an unsupervised deep learning atlas-registration method to align their fields of view (FOVs). For this we employ two ConvNets: one for atlas-registration and one for calcium scoring, as shown in Figure~\ref{fig:pipeline}. The atlas-registration ConvNet makes the FOV of input CT images alike using Deep Learning Image Registration (DLIR)~\cite{devos2017registration, devos2018media} further developed to facilitate atlas-registration.
Subsequently, a calcium scoring ConvNet predicts the calcium score in image slices mimicking clinical calcium scoring with the Agatston score. When desired, decision feedback can be queried for every slice with a predicted calcium score. For this purpose, a visual attention heatmap accurately reveals the regions that contributed to the calcium score. The method provides robust and accurate predictions of calcium scores and it is computationally efficient, obtaining an Agatston score in less than 0.3\,s in cardiac and chest CT.

\section{Data}
This study included two datasets used in previous studies that presented automatic coronary calcium scoring in cardiac CT~\cite{wolterink2015} and in chest CT~\cite{lessmann2018}. To allow a direct comparison of methods, the original training, validation, and test set distributions were used.

\subsection{Cardiac CT}
The set of 903 cardiac CT scans (age range: 18 to 88 years, 31\% women) originates from a set of routinely acquired scans for clinical calcium scoring of the University Medical Center Utrecht, Utrecht, The Netherlands. The need for informed consent was waived by the local Medical Research Ethics Committee. Scans were acquired with a 256-detector row Philips Brilliance iCT scanner (tube voltage 120\,kVp, tube current 55\,mAs) during a single breath-hold, with ECG-triggering and without contrast enhancement. The images were reconstructed to 3\,mm slice thickness and slice increment with in-plane resolution ranging from 0.29\,mm to 0.49\,mm, depending on patient size. The dataset was divided into 237 scans for training, 136 scans for validation, and 530 scans were in the hold-out test set only used for final evaluation.

\subsection{Chest CT}
The set of 1,687 chest CT scans (age range: 43 to 74 years, 39\% women) originates from a set of 6,000 available baseline scans from the National Lung Screening Trial (NLST)~\cite{nlst2011b}. All scans were acquired during inspiratory breath-hold without contrast enhancement. Scans were acquired in 31 different hospitals with 120 or 140\,kVp tube voltage and 30-160\,mAs tube current. Axial images slices were reconstructed with varying kernels, varying slice thickness (1.00-3.00\,mm), varying slice increments (0.63-3.00\,mm), and with varying in-plane resolutions (0.49-0.98\,mm per voxel). In our study, scans with less than 100 slices or slices thicker than 3.00\,mm were not considered, because they were not adequate for calcium scoring. Furthermore, the scans were resampled to 3.00\,mm slice thickness and 1.50\,mm slice increment to make the scans suitable for calcium scoring \cite{rutten2011}. The dataset was divided into 1,012 scans for training, 169 scans for validation, and 506 scans were in the hold-out test set only used for final evaluation.

\begin{table}
\color{black}
    \centering
    \caption{Number of scans per CVD risk category for training, validation, and test sets. CVD risk categorization is based on the total Agatston score per scan: \rom{1}: very low $<1$, \rom{2}: low $[1, 10)$, \rom{3}: moderate $[10, 100)$, \rom{4}: moderately high $[100, 400)$, \rom{5}: high $\geq400$}
    \label{tab:riskcatsalldata}
    \begin{tabular}{cc|ccccc}
                    & &  \rom{1} & \rom{2} & \rom{3} & \rom{4} & \rom{5}\\
                    \hline
         \multirow{3}{*}{Cardiac CT} & Training & 120 & 14 & 33 &  29 &  41  \\
          & Validation & 68 & 14 & 28 &  15 &  11  \\
          & Test & 260 & 49 & 89 &  70 &  62  \\
          \hline
         \multirow{3}{*}{Chest CT}  & Training & 272 & 76 & 207 & 205 & 252 \\
          & Validation & 39 & 14 & 46 & 30 & 40 \\
           & Test & 128 & 42 & 99 & 112 & 125 \\
    \end{tabular}
\end{table}

\subsection{Reference standard}
\label{sec:refstandard}
The reference standard was defined by experts who manually identified CAC lesions in the scans. CAC lesions were segmented following a standard procedure: region growing was used to select 26-connected voxels $\geq$130\,HU. In the chest CTs with low radiation dose this procedure could lead to faulty segmentations (i.e. leakage) because of excessive noise. In such cases annotations were manually corrected by voxel painting~\cite{lessmann2018}.
Agatston scores were calculated in each axial slice for training. Total Agatston scores for each scan were calculated for final evaluation. Additionally, each subject was assigned to one of five CVD risk categories~\cite{rumberger1999} based on the Agatston score: very low: $<$1; low: [1, 10), moderate: [10, 100), high: [100, 400), very high: $\geq$400. Table~\ref{tab:riskcatsalldata} provides an overview of the number of scans per risk category per dataset.

\section{Methods}
\begin{figure}
    \centering
    \includegraphics[width=\linewidth]{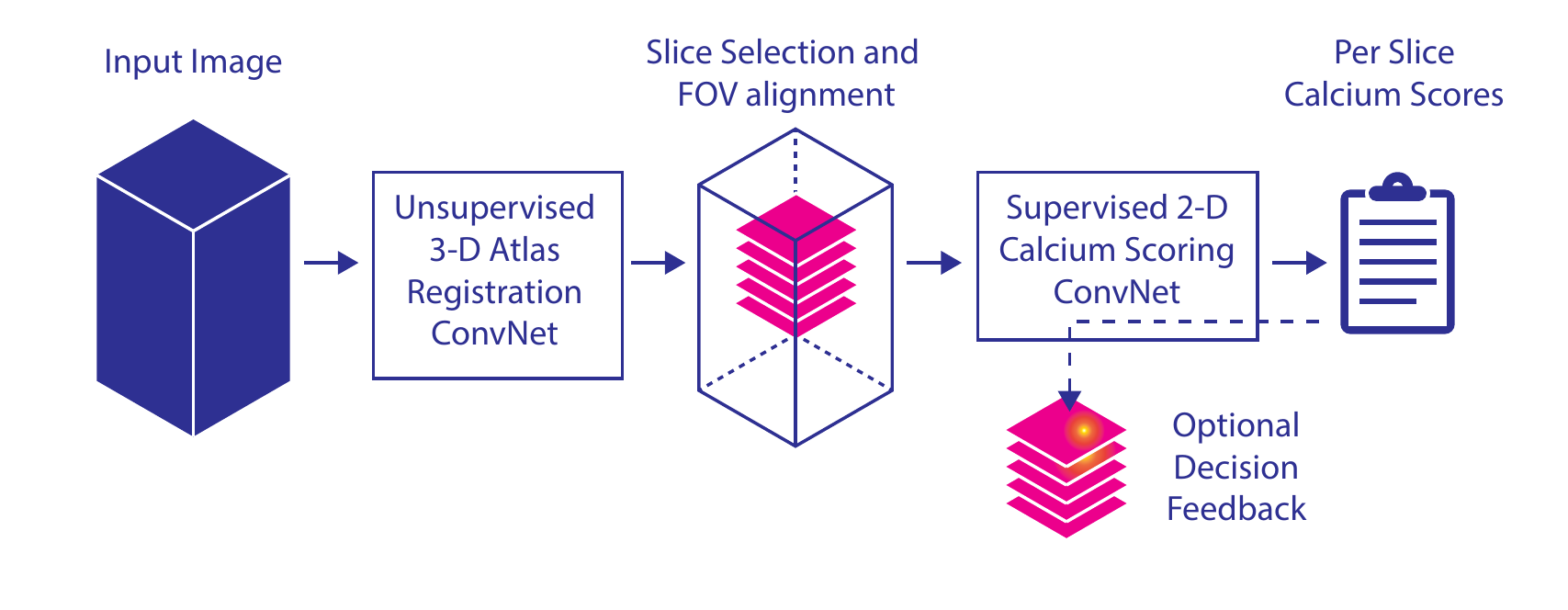}
    \caption{Schematics of the proposed method. Input CTs of varying FOV are first aligned using an atlas-registration ConvNet. Subsequently, a calcium scoring ConvNet is used for direct calcium scoring in image slices. Finally, decision feedback can be visualized when desired.}
    \label{fig:pipeline}
\end{figure}

The method employs two ConvNets in sequence (Figure~\ref{fig:pipeline}). The first ConvNet registers input CTs to an cardiac CT atlas-image. The second ConvNet performs calcium scoring. When desired, visual feedback can be queried for image slices with a score. For this purpose an attention heatmap reveals the regions that contributed to the calcium score.

\subsection{Atlas-registration strategy}
\label{sec:method:registration}
\begin{figure}
    \centering
    \includegraphics[width=\linewidth]{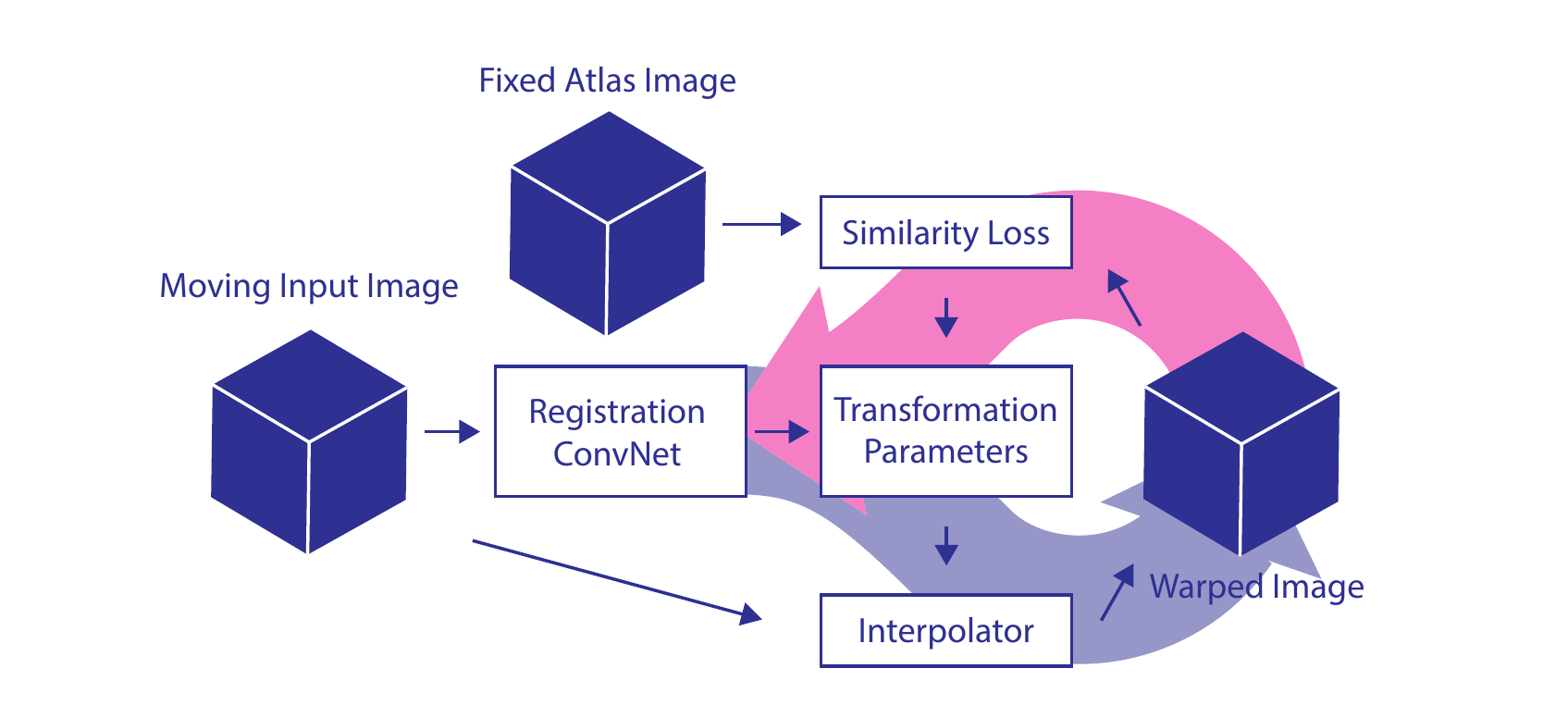}
    \caption{DLIR framework used to train a registration ConvNet. During a forward pass (indicated by the thick blue arrow) the registration ConvNet analyzes moving images and outputs transformation parameters. The transformation parameters are used by the interpolator to warp the moving image. During a backward pass (indicated by the thick red arrow) an image similarity loss (i.e. dissimilarity) is determined between the warped image and a fixed template image, and the resulting loss is backpropagated trough the ConvNet. The ConvNet is trained in multiple iterations of forward and backward passes, with mini-batch stochastic gradient descent. Once the ConvNet has been trained for registration it can take a moving image as its input and it can output registration parameters in one pass, thus non-iteratively.}
    \label{fig:dlirframework}
\end{figure}

An atlas-registration ConvNet ensures that all input images have a similar FOV and resemble a cardiac CT.
The ConvNet is trained with a modified version of our framework for Deep Learning Image Registration (DLIR)~\cite{devos2017registration}. The DLIR framework uses an end-to-end unsupervised approach that trains a ConvNet for image registration. Similar to a conventional intensity-based image registration framework it exploits optimization of an image similarity metric.
Figure~\ref{fig:dlirframework} shows the schematics of training an atlas-registration ConvNet using the atlas image as a static fixed image. The task of the ConvNet is to analyze moving images and predict the transformation parameters that warp the moving images to the atlas-image. Image similarity between the atlas and the warped image, is used for backpropagation during training. By optimizing image similarity (e.g. minimizing negative cross correlation) with gradient descent, the atlas-registration ConvNet learns the registration task in an unsupervised manner. After training, the ConvNet can register unseen moving images in one shot.

A cardiac atlas-image is created using an iterative inter-subject registration strategy~\cite{jongen2004}. With this strategy an initial atlas image is made by averaging multiple images. The atlas image is iteratively refined by registering the individual images to the atlas. Subsequently, the final atlas image is used to train the atlas-registration ConvNets for cardiac and chest CT alignment used for subsequent calcium scoring.

\subsection{Atlas-registration ConvNet training}
\begin{figure}
    \centering
    \begin{tikzpicture}
        \node (img) at (0,0) {\includegraphics[width=.35\linewidth]{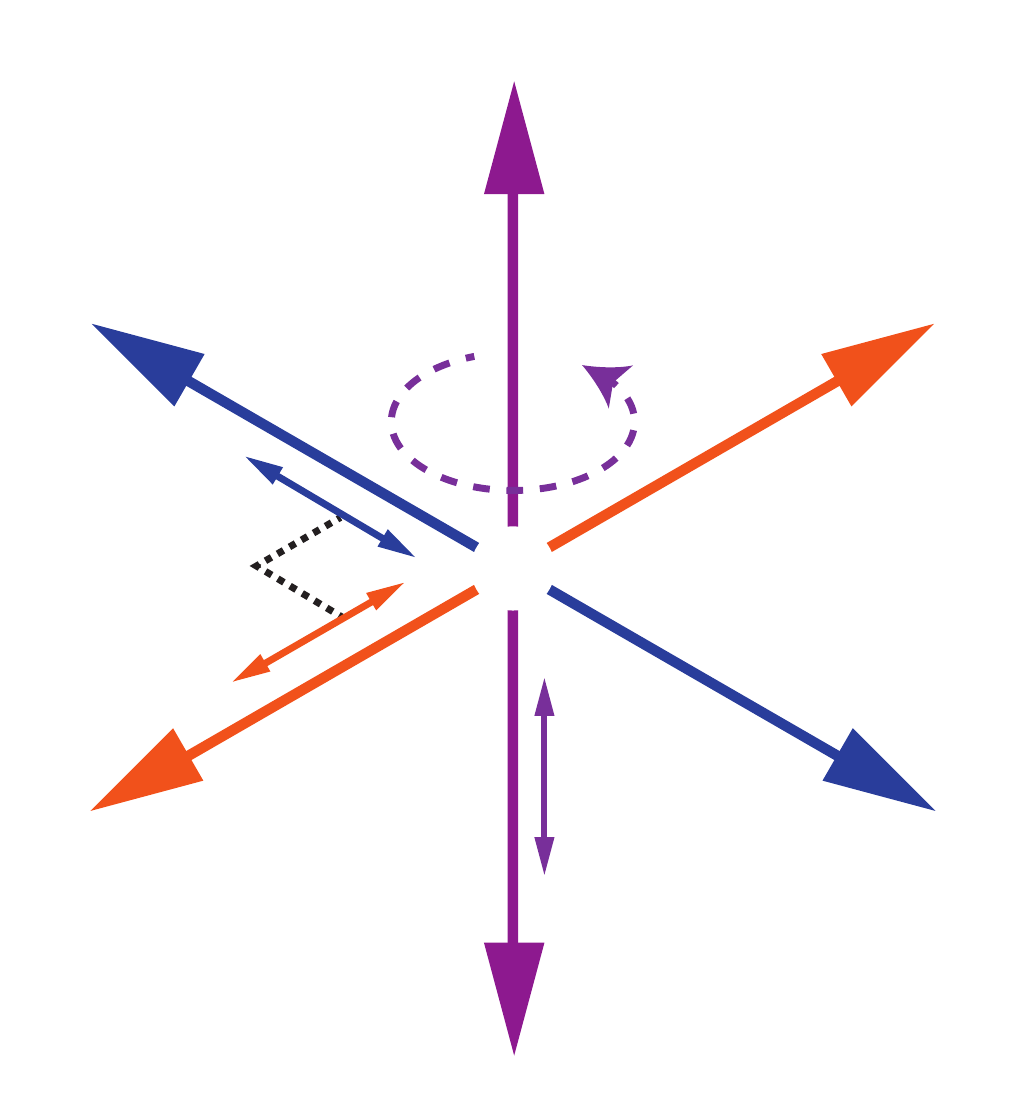}};
        \node at (32pt, 26pt) {$t_y$};
        \node at (38pt,-14pt) {$t_x$};
        \node at (-7pt,41pt) {$t_z$};
        \node at (-32pt,-2pt) {$s_{xy}$};
        \node at (10pt,-20pt) {$s_z$};
        \node at (15pt,20pt) {$\theta_z$};
    \end{tikzpicture}
    \caption{Rigid transformation model used for to train the registration ConvNet. The six degrees of freedom allow translation in any direction, rotation around the axial axis, and uniform scaling in the axial plane independent from scaling along the axial direction. By constraining the registration to the proposed transformation model, we can trivially exploit the model parameters for selection and warping of axial slices that are presented to the calcium scoring ConvNet.}
    \label{fig:degrees_of_freedom}
\end{figure}
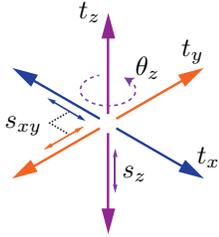

For registration we propose a global 3-D rigid registration model with six degrees of freedom (shown in Figure~\ref{fig:degrees_of_freedom}). The model allows translations in any direction, but rotations are restricted to the axial ($z$) axis. Furthermore, scaling in the axial plane is isotropic and independent from scaling along the axial axis. These restrictions preserve the relation of reference Agatston scores that are defined on the original (unregistered) axial image slices. This facilitates training of the subsequent calcium scoring ConvNet.

We use a computationally efficient ConvNet architecture that is listed in Table~\ref{tab:convnetdesigns}. For fast analysis, images are downsampled close to 3\,mm isotropic voxel dimensions; i.e. 6$\times$6$\times$1 downsampling for cardiac CT, and 6$\times$6$\times$2 downsampling for chest CT using average pooling. 
The ConvNet has three alternating layers of 3$\times$3$\times$3 convolutions and 2$\times$2$\times$2 average pooling and those are followed by two layers of 3$\times$3$\times$3 convolution. To facilitate a fixed output, global average pooling is applied before connection with two fully connected layers. The final output layer has six nodes, one for each transformation parameter. Throughout the network exponential linear units are used for activation, except in the output nodes. Three output nodes are unconstrained translation parameters ($t_x$, $t_y$, $t_z$), the rotation parameter ($\theta_z$) is constrained with a hyperbolic tangent between $-\pi$ and $\pi$, and the two scaling parameters ($s_{xy}$, $s_z$) are constrained with a hyperbolic tangent between $0.25$ and $4$ scaling factors. These output parameters are used to constitute the following 3-D transformation matrix:
\begingroup % keep the change local
\begin{equation*}
    T_\textrm{3D} =
    \setlength\arraycolsep{2pt}
    \begin{bmatrix}
        1   &   0   &   0   &   t_x \\
        0   &   1   &   0   &   t_y \\
        0   &   0   &   1   &   t_z \\
        0   &   0   &   0   &   1   \\
    \end{bmatrix}
    \begin{bmatrix}
        \cos\theta_z   &   -\sin\theta_z   &   0   &   0   \\
        \sin\theta_z   &   \phantom{-}\cos\theta_z   &   0   &   0   \\
        0   &   0   &   1   &   0   \\
        0   &   0   &   0   &   1   \\
    \end{bmatrix} 
    \begin{bmatrix}
        s_{xy}  &   0      &   0   &   0 \\
        0       &   s_{xy} &   0   &   0 \\
        0       &   0      &   s_z &   0 \\
        0       &   0      &   0   &   1   \\
    \end{bmatrix}
\end{equation*}
\endgroup

\subsection{Atlas-registration ConvNet inference}
We train an atlas-registration ConvNet for 3-D registration, but we use it for slice selection and 2-D warping. As a consequence, correspondence is guaranteed between warped axial slices and the per-slice calcium scores.
Axial image slices are extracted from the original image from $t_z$ to $t_z + d_z/s_z$, where $d_z$ is the depth of the atlas image along the axial axis.
These slices are resampled using bi-linear interpolation to a 256$\times$256 grid with the following 2-D transformation matrix:
\begin{equation*}
    T_\textrm{2D} = 
    \begin{bmatrix*}
        1   &   0   &   t_x \\
        0   &   1   &   t_y \\
        0   &   0   &   1   \\
    \end{bmatrix*}
    \begin{bmatrix*}
        \cos\theta_z   &   -\sin\theta_z  &   0 \\
        \sin\theta_z   &   \phantom{-}\cos\theta_z   &   0 \\
        0   &   0               &   1 \\
    \end{bmatrix*}
    \begin{bmatrix*}
        s_{xy}  &   0      &   0 \\
        0       &   s_{xy} &   0 \\
        0       &   0      &   1 \\
    \end{bmatrix*}
\end{equation*}

\begin{table}
\centering
\caption{Efficient ConvNet architectures were used for atlas-registration as well as calcium scoring.}
\label{tab:convnetdesigns}
\begin{tabular}{ll}
                    \textbf{Atlas-Registration ConvNet}   & \textbf{Calcium Scoring ConvNet}     \\
                   \hline
                    512$\times$512$\times$N 3-D input  & 256$\times$256 2-D input \\
                    6$\times$6$\times$\{1,2\} Avg. Pooling  & 224$\times$224 cropping \\
                    \hline
                    32*3$\times$3$\times$3 Convolutions       & 32*3$\times$3 Convolutions        \\
                    \phantom{32*}2$\times$2$\times$2 Avg. Pooling          & \phantom{32*}2$\times$2 Max Pooling            \\
                   32*3$\times$3$\times$3 Convolutions       & 32*3$\times$3 Convolutions        \\
                    \phantom{32*}2$\times$2$\times$2 Avg. Pooling           & \phantom{32*}2$\times$2 Max Pooling            \\
                    32*3$\times$3$\times$3 Convolutions       & 32*3$\times$3 Convolutions        \\
                    \phantom{32*}2$\times$2$\times$2 Avg. Pooling               & \phantom{32*}2$\times$2 Max Pooling            \\
                   32*3$\times$3$\times$3 Convolutions       & 32*3$\times$3 Convolutions        \\
                    32*3$\times$3$\times$3 Convolutions                           & \phantom{32*}2$\times$2 Max Pooling            \\
                    Global Avg. Pooling       & 32*3$\times$3 Convolutions        \\
                                                & \phantom{32*}2$\times$2 Max Pooling            \\
                                               & 32*3$\times$3 Convolutions        \\
                          & \phantom{32*}2$\times$2 Max Pooling            \\
                   \hline
                    64 Fully Connected Nodes    & 64 Fully Connected Nodes   \\
                    64 Fully Connected Nodes    & 64 Fully Connected Nodes   \\
                   \phantom{6}6 Output Nodes     & \phantom{6}1 Output Node     \\
\end{tabular}
\end{table}

\subsection{Calcium scoring ConvNet}
\label{sec:cacscoremethod}
The calcium scoring ConvNet employs direct regression to predict an Agatston score from input axial image slices. The choice of 2-D ConvNets, in favor of 3-D ConvNets, is based on the number of samples that are available for training. There are more image slices available than image volumes. Furthermore, {2-D} image analysis mimics clinical calculation of the Agatston calcium score that is performed in {2-D} axial slices: 
\begin{equation*}
\textrm{Agatston Score} = {\sum_{S \in V}\sum_{l \in S}{A_l w_l\frac{i_S}{t_S}}}\,.
\end{equation*}
where $l$ is a 2-D CAC lesion in a slice $S$ of a CT volume $V$. $A_l$ is the area of the lesion. The weighted intensity $w_l$ is based on the maximum radio-density in HU of a 2-D lesion in the following manner: 1 = [130, 200), 2 = [200, 300), 3 = [300, 400), and 4 = $\geq$400. The Agatston score is corrected when image slices are overlapping, thus when slice increment $i_S$ is not equal to slice thickness $t_S$~\cite{ohnesorg2002}.

Agatston scores are dependent on the CAC lesion area. Given that input images have different voxels sizes, we chose to simplify the prediction task by determining a pseudo-Agatston score. This score is obtained by cancelling out the axial pixel dimensions, the slice increment, and the slice thickness of the original Agatston score. The resulting target is the product of the number of voxels in a lesion $n_l$, the predicted slice scaling factor $s_{xy}$, and the weighted intensity~$w_l$:
\begin{equation*}
\textrm{Pseudo-Agatston Score} = \sum_{l \in S} n_l  \cdot s_{xy} \cdot w_l\,.
\end{equation*}
The calcium scoring ConvNet uses an efficient architecture that is listed in Table~\ref{tab:convnetdesigns}. It analyzes random image croppings of 224$\times$224 pixels during training and center croppings during application. It has alternating layers of 3$\times$3 convolutions and $2\times2$ max pooling, followed by two fully connected layers, and an output layer of one node. Throughout the network batch normalization~\cite{ioffe2015} is used and exponential linear units are used for activation\cite{clevert2016}. The final output node has a linear output to facilitate continuous prediction. However, given that clinically used CVD risk categories are exponentially increasing, the task of the calcium scoring ConvNet was modified to learn a log-transform of the pseudo-Agatston score:
\begin{equation*}
L = |\hat{y} - \ln(y + 1)|\,,
\end{equation*}
where $\hat{y}$ is the predicted score, and $y$ is the reference pseudo-Agatston score. The log-transform induces relatively high penalties for erroneous low calcium score predictions, and relatively low penalties for erroneous high calcium score predictions. Consequently, higher precision is forced for lower calcium burden, which is favorable for CVD risk stratification. During application of the calcium scoring ConvNet, the predicted outputs are converted to the original Agatston scores.

\subsection{Decision feedback}
By employing regression of calcium scores, we circumvent time-consuming intermediate segmentation. On the other hand, it may be desirable to visualize regions in image slice that contributed to the calcium score. Inspired by the study of Zeiler and Fergus~\cite{zeilerfergus2014}, we provide such visualization by using a deConvNet. The deConvNet uses the same operations of filtering and pooling as a ConvNet, but in reverse order from output to input. The reverse operations map the activities back to the input pixel space, and it shows which input patterns originally contributed to the activations in the feature maps. To obtain a smooth visual attention heatmap, the deConvNet is applied until the third convolutional layer, by taking the absolute value per feature of this layer, and by summing these features along the feature map dimension to get 2-D matrix. Using third order interpolation we obtain a smooth map that can be superpositioned on the image slice as a heatmap. This resulting heatmap visualizes attention by highlighting the regions that contributed to the Agatston score.

\section{Evaluation}
Automatically predicted per-subject Agatston scores were compared with manually determined reference scores. Evaluations were performed on the hold-out test sets which were not used during method development. Two-way mixed intra-class correlation coefficient (ICC) for absolute agreement was computed and Bland-Altman analysis was performed to evaluate bias between predicted and reference Agatston scores. 
In addition, for each subject, CVD risk category was determined based on the Agatston score as defined in section~\ref{sec:refstandard}.
% score~\cite{rumberger1999}: very low ($<$1), low (1--10), moderate (10--100), moderately high (100--400), and high ($\geq$400).
Agreement between predicted and reference CVD risk categories was determined using accuracy and Cohen's linearly weighted kappa ($\kappa$).

\section{Experiments and results}
In this section we evaluate the atlas-registration ConvNet, the calcium scoring ConvNet, and the quality of decision feedback. In addition, we will evaluate whether the calcium scoring ConvNet requires to be trained on all data, or whether it can be trained on one dataset and applied to the other. Finally, we will compare state-of-the-art automatic calcium scoring methods with the proposed method. All experiments were performed with Theano~\cite{theano2016}, Lasagne~\cite{lasagne2015}, and OpenCV~\cite{opencv} on an Intel Xeon E5-1620 3.60\,GHz CPU with an NVIDIA Titan X GPU. 

\subsection{Atlas-registration ConvNet}
\begin{figure}
    \centering
    \begin{tabular}{p{.1em}cp{.1em}cc}
        &\footnotesize{Training Set} & &\multicolumn{2}{c}{\footnotesize{Test Set}}\\
         &\phantom{0}\footnotesize{Cardiac CT Atlas} \vspace{-1mm} & &\footnotesize{Cardiac CTs}  \vspace{-1mm}& \footnotesize{Chest CTs} \vspace{-1mm}\\
    \rotatebox[origin=c]{90}{\footnotesize{Initial Atlas}}&
    \subfloat[]{%
    \vspace{-\parskip}
    \begin{minipage}[t]{0.25\linewidth}
        \centering
        \includegraphics[width=\linewidth]{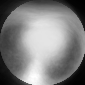}\vspace{1mm}
        \includegraphics[width=\linewidth]{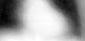}\vspace{1mm}
        \includegraphics[width=\linewidth]{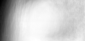}\vspace{-3mm}
        \label{fig:atlas:atlas1}
    \end{minipage}}
    &
    \rotatebox[origin=c]{90}{\footnotesize{Before Registration}}
    &
    \subfloat[]{\begin{minipage}[t]{0.25\linewidth}
        \centering
        \includegraphics[width=\linewidth]{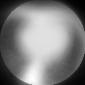}\vspace{1mm}
        \includegraphics[width=\linewidth]{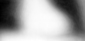}\vspace{1mm}
        \includegraphics[width=\linewidth]{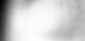}\vspace{-3mm}
        \label{fig:atlas:cardiac1}
    \end{minipage}}
    &
    \subfloat[]{\begin{minipage}[t]{0.25\linewidth}
        \centering
        \includegraphics[width=\linewidth]{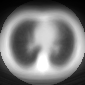}\vspace{1mm}
        \includegraphics[width=\linewidth]{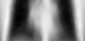}\vspace{1mm}
        \includegraphics[width=\linewidth]{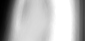}\vspace{-3mm}
        \label{fig:atlas:chest1}
    \end{minipage}}\\
    \rotatebox[origin=c]{90}{\footnotesize{Refined Atlas}}
    &
    \subfloat[]{\begin{minipage}[t]{0.25\linewidth}
        \centering
        \includegraphics[width=\linewidth]{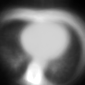}\vspace{1mm}
        \includegraphics[width=\linewidth]{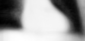}\vspace{1mm}
        \includegraphics[width=\linewidth]{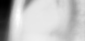}\vspace{-3mm}
        \label{fig:atlas:atlas2}
    \end{minipage}}
    &
    \rotatebox[origin=c]{90}{\footnotesize{After Registration}}&
    \subfloat[]{\begin{minipage}[t]{0.25\linewidth}
        \centering
        \includegraphics[width=\linewidth]{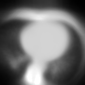}\vspace{1mm}
        \includegraphics[width=\linewidth]{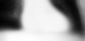}\vspace{1mm}
        \includegraphics[width=\linewidth]{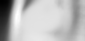}\vspace{-3mm}
        \label{fig:atlas:cardiac2}
    \end{minipage}}
    &
    \subfloat[]{\begin{minipage}[t]{0.25\linewidth}
        \centering
        \includegraphics[width=\linewidth]{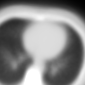}\vspace{1mm}
        \includegraphics[width=\linewidth]{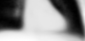}\vspace{1mm}
        \includegraphics[width=\linewidth]{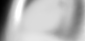}\vspace{-3mm}
        \label{fig:atlas:chest2}
    \end{minipage}}\\
    \end{tabular}
    \caption{Cross-sectional views of the generated atlas images and the average images that illustrate registration performance. An initial atlas image \protect\subref{fig:atlas:atlas1} was made from 237 cardiac CTs that were aligned with their geometric centers. This atlas image was used to train an atlas-registration ConvNet to obtain a refined atlas image \protect\subref{fig:atlas:atlas2}. Finally, this refined atlas was used to train the atlas-registration ConvNets for cardiac and chest CT FOV alignment. To illustrate the performance of the resulting ConvNets we provide average images of the cardiac CT and chest CT test sets before registration in \protect\subref{fig:atlas:cardiac1} and \protect\subref{fig:atlas:chest1}, and after registration \protect\subref{fig:atlas:cardiac2} and \protect\subref{fig:atlas:chest2}. For each example we show from top to bottom center slices of axial, coronal, and sagittal views.}
    \label{fig:atlas}
\end{figure}

Figure~\ref{fig:atlas:atlas1} shows the initial atlas image that was created by aligning all cardiac training images using their geometric centroids. We chose the median dimensions and voxels sizes of all the cardiac training images define the atlas image space.
The atlas can be iteratively refined, but given the constraints of the global registration model used here, only one update was sufficient. The final atlas image, shown in Figure~\ref{fig:atlas:atlas2}, was used to train the atlas-registration ConvNets for cardiac and chest CT alignment. Thus, in total three ConvNet instances were trained: one to create an atlas image, one for cardiac CT alignment, and one for chest CT alignment. All ConvNets were trained in 15,000 iterations with mini-batches containing 32 randomly selected images. Training took about 40 hours per ConvNet. Adam~\cite{kingma2014} was used with a learning rate of 0.001 for mini-batch gradient descent. To illustrate performance of the atlas-registration ConvNets, Figure~\ref{fig:atlas} shows images before and after registration. Figure~\ref{fig:atlas:cardiac1} shows the average image of the 530 cardiac CT images from the test set before registration and Figure~\ref{fig:atlas:cardiac2} shows these images after registration. Similarly, Figure~\ref{fig:atlas:chest1} shows an average image of the 506 chest CTs before registration and Figure~\ref{fig:atlas:chest2} shows these after registration. Note the similarity of the registered image with the refined atlas image shown in Figure~\ref{fig:atlas:atlas2}.

Quantitative evaluation of registration results revealed that registration erroneously cropped CAC out of the selected slices. Between one and four image slices containing CAC were not selected in three cardiac CTs and three chest CTs. Upon closer inspection, two of the chest CTs had calcifications in the aortic arch and descending aorta incorrectly labeled as CAC in the reference, thereby affecting CVD risk categorization. Nevertheless, these annotations were left uncorrected in further analysis to facilitate a fair comparison with previously developed methods. The registration errors did not have an adverse effect on CVD risk categorization in the other cases.

\subsection{Calcium scoring ConvNet}
\label{sec:calciumscoringconvnet}
The calcium scoring ConvNet was trained in 150,000 iterations using Adam~\cite{kingma2014}. Training took 21 hours with 100 image slices per mini-batch randomly selected from the registered image slices taken from the cardiac and chest CT training sets. High imbalance between the minority of slices with a calcium score and the majority of slices with zero calcium score prevented convergence during ConvNet training. To ensure convergence, the amount of image slices with CAC (Agatston score~$>0$) and without CAC (Agatston score~$=0$) were balanced during training. To prevent bias, training continued on the full imbalanced training set after 10,000 iterations. Additionally, we ensured stable convergence by decreasing the learning rate to 10\% of its previous value every 50,000 iterations.

\begin{figure}
    \centering
    \subfloat[Cardiac CT]{\begin{minipage}[t]{.9\linewidth}
        \includegraphics[width=\linewidth,trim={.5cm .0cm 1cm .5cm},clip]{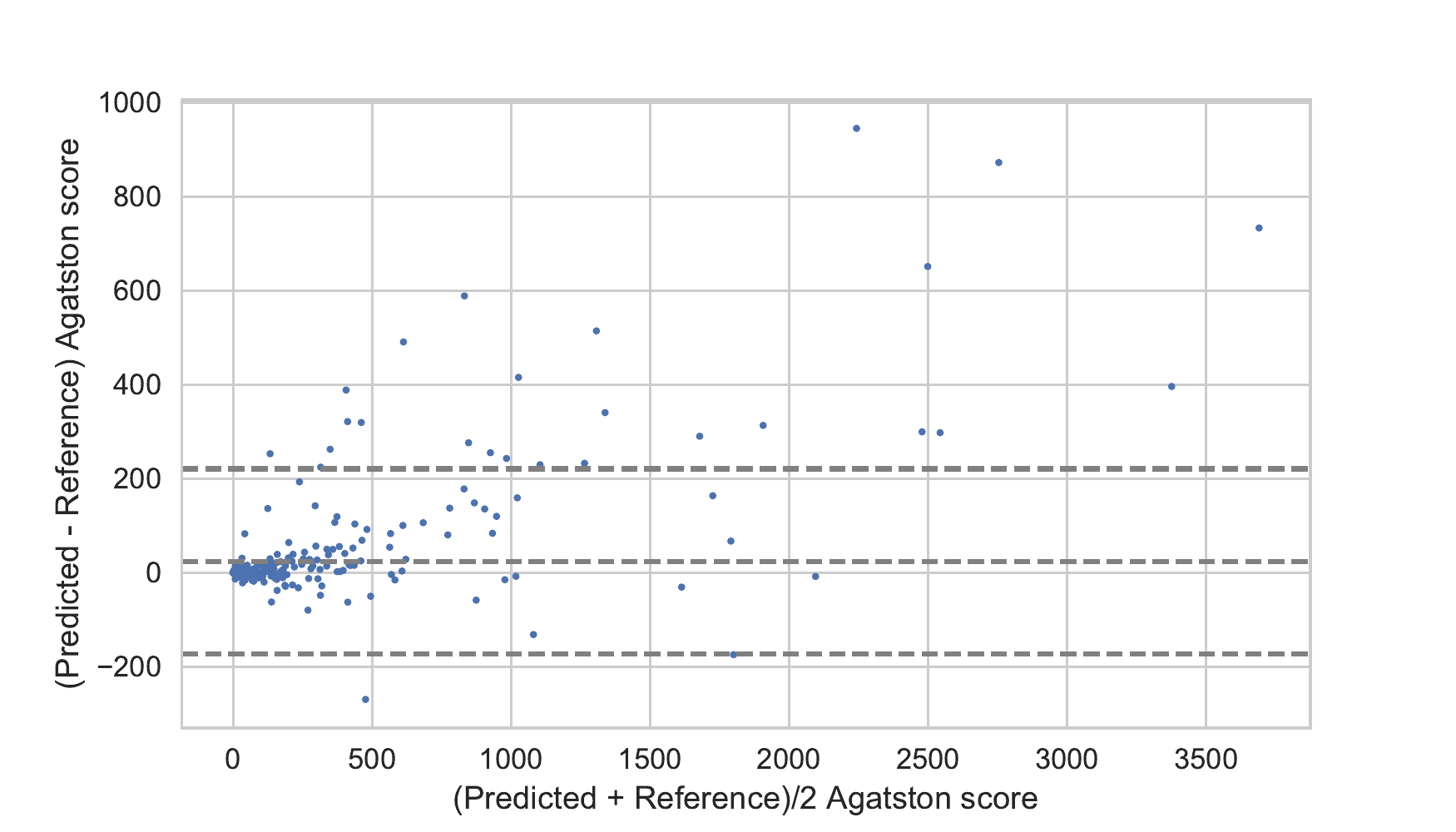}
        \label{fig:blandaltman:cardiac}
    \end{minipage}}\\
    \subfloat[Chest CT]{\begin{minipage}[t]{.9\linewidth}
        \includegraphics[width=\linewidth,trim={.5cm .0cm 1cm .5cm},clip]{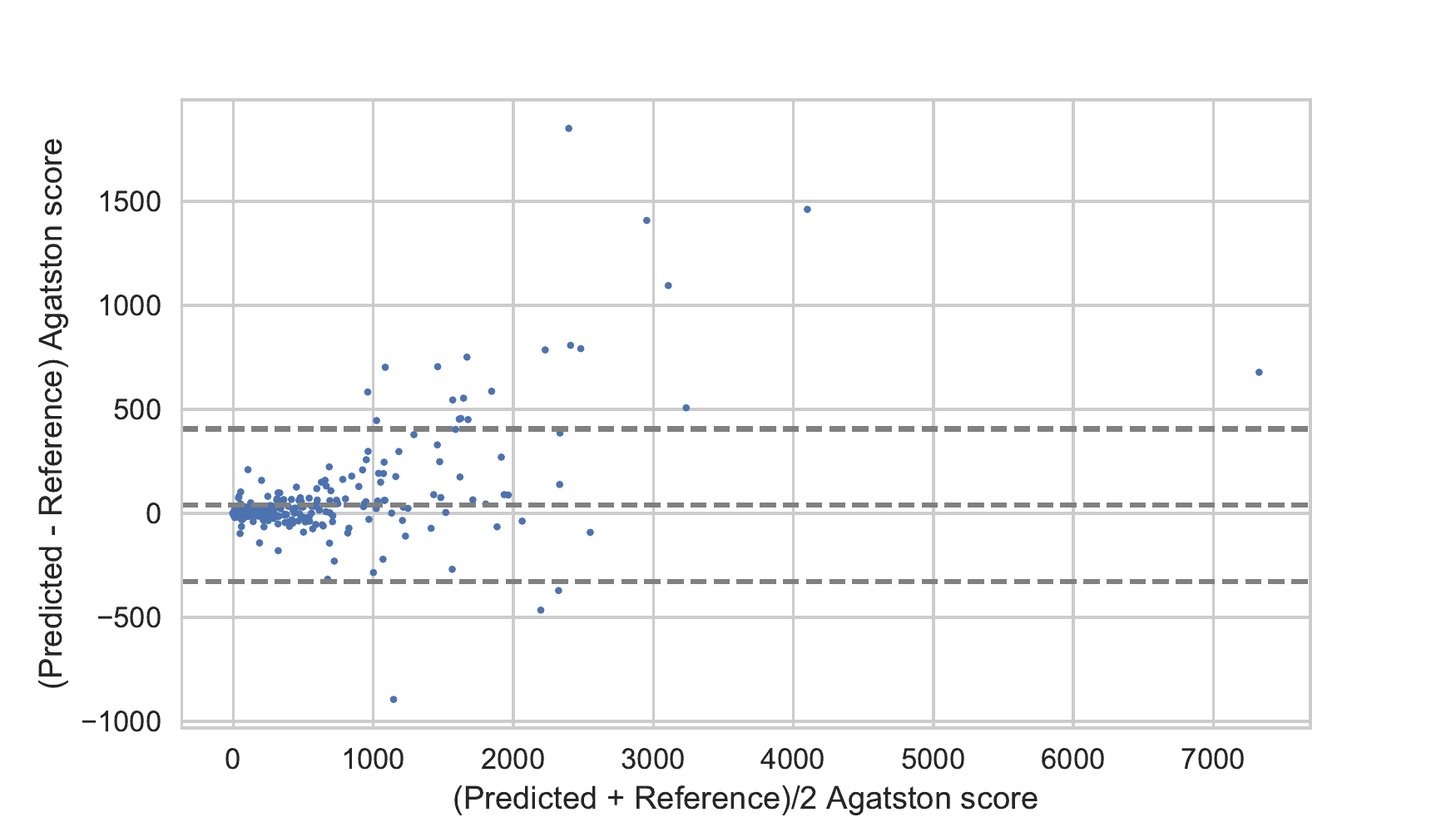}
        \label{fig:blandaltman:chest}
    \end{minipage}}
\caption{Bland-Altman plots showing agreement between predicted and reference per subject Agatston scores in the cardiac CT~\protect\subref{fig:blandaltman:cardiac} and chest CT~\protect\subref{fig:blandaltman:chest} datasets. Limits of agreement are $\pm$1.96 SD, the positive biases in both datasets are mainly caused by overestimations of the higher Agatston scores.}
    \label{fig:blandaltman}
\end{figure}

\begingroup
\begin{table}
        \centering
        \caption{Confusion matrices showing agreement in CVD risk categorization based on the total Agatston scores: \rom{1}: very low $<1$, \rom{2}: low $[1, 10)$, \rom{3}: moderate $[10, 100)$, \rom{4}: moderately high $[100, 400)$, \rom{5}: high $\geq400$. The method is evaluated separately on the test sets of cardiac CTs (left) and chest CTs (right). The corresponding linearly weighted $\kappa$ is shown below the confusion matrices.}
        \label{fig:confusion:fullset}
        \setlength\tabcolsep{4pt}
        \begin{tabular}{rr|ccccc}
        \multicolumn{2}{c}{}&\multicolumn{5}{c}{\textit{Predicted}}\\
        {} &{} &    \rom{1} &  \rom{2} &  \rom{3} &  \rom{4} &   \rom{5} \\
        \cline{2-7}
        \multirow{5}{*}{\rotatebox[origin=c]{90}{\textit{Reference}}}
        &\rom{1} &      \textbf{259} &       0 &        1 &        0 &        0 \\
        &\rom{2} &       9 &       \textbf{36} &       4 &        0 &        0 \\
        &\rom{3} &        2 &       3 &       \textbf{82} &        2 &        0 \\
        &\rom{4} &        0 &        1 &       2 &       \textbf{65} &        2 \\
        &\rom{5} &        0 &        0 &        0 &       11 &       \textbf{51} \\
        \multicolumn{2}{c}{}&\multicolumn{5}{c}{\textit{Cardiac CT} $\kappa=0.95$}
        \end{tabular}
        \begin{tabular}{rr|ccccc}
        \multicolumn{2}{c}{}&\multicolumn{5}{c}{\textit{Predicted}}\\
        {} &{} &    \rom{1} &  \rom{2} &  \rom{3} &  \rom{4} &   \rom{5} \\
        \cline{2-7}
        \multirow{5}{*}{\rotatebox[origin=c]{90}{\textit{Reference}}}
        &\rom{1} &       \textbf{118} &       6 &        4 &        0 &        0 \\
        &\rom{2} &       8 &       \textbf{29} &        5 &        0 &        0 \\
        &\rom{3} &        3 &       8 &       \textbf{85} &        3 &        0 \\
        &\rom{4} &        1 &        1 &       7 &       \textbf{99} &        4 \\
        &\rom{5} &        0 &        0 &        0 &       3 &       \textbf{122} \\
        \multicolumn{2}{c}{}&\multicolumn{5}{c}{\textit{Chest CT} $\kappa=0.93$}
        \end{tabular}
\end{table}
\endgroup

After training, the test sets were used to evaluate the calcium scoring ConvNet. Per-subject scores show high intraclass correlation coefficients (ICC); the ICC for cardiac CT and chest CT were both 0.98 with 95\% confidence intervals of 0.98 to 0.99. Slight positive bias in cardiac and chest CT is visualized with the Bland-Altman plots shown in Figure~\ref{fig:blandaltman}. This was mainly caused by overestimations of the higher Agatston scores. However, this was not noticeable in CVD risk stratification. Table~\ref{fig:confusion:fullset} shows confusion matrices of predicted risk categories vs. the manual reference standard. In cardiac CT calcium scoring only four scans were two categories off, and in chest CT calcium scoring eight scans were two categories off. The scan that was three categories off was a scan with incorrectly annotated aorta calcium, as discussed in the previous section. Nonetheless, overall agreement was \textit{almost perfect}~\cite{mchugh2012} with Cohen's linearly weighted $\kappa$s of 0.95 in cardiac CT and 0.93 in chest CT. Accuracy in CVD risk categorization was 0.93 for cardiac CT and 0.90 for chest CT. Because efficient network architectures are used, the method is able to achieve high speed when used on a single CPU core: within 5\,s a score for cardiac CT is obtained and within 11\,s a score for chest CT is obtained. When using a GPU, calcium scoring can be performed in real-time. Including image registration and image resampling, a calcium score for cardiac CT is obtained in less than 0.15\,s and for chest CT in less than 0.30\,s. 

\subsection{Decision feedback}
\begin{figure*}[]
    \centering
    \subfloat[Predicted: 12 -- Reference: 12]{\label{fig:attention:cardiac1}\includegraphics[height=3.2cm,trim={2.2cm 1cm 3cm 0cm},clip]{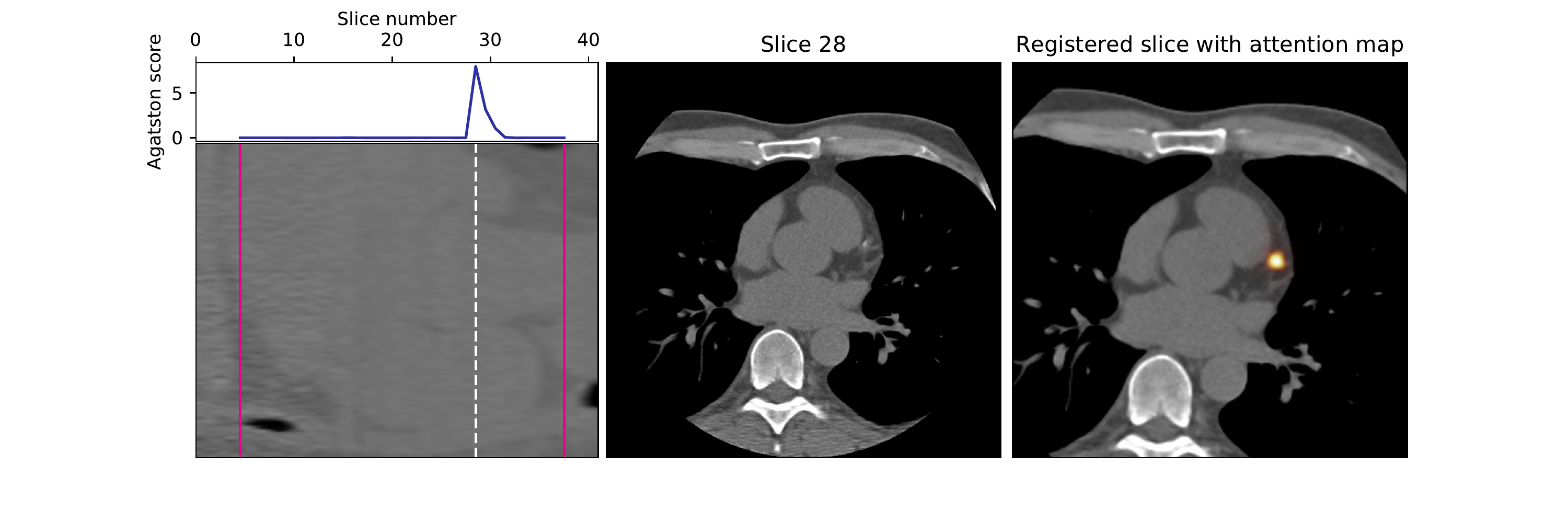}}
    \subfloat[Predicted: 124 -- Reference: 124]{\label{fig:attention:chest1}\includegraphics[height=3.2cm,trim={2.2cm 1cm 3cm 0cm},clip]{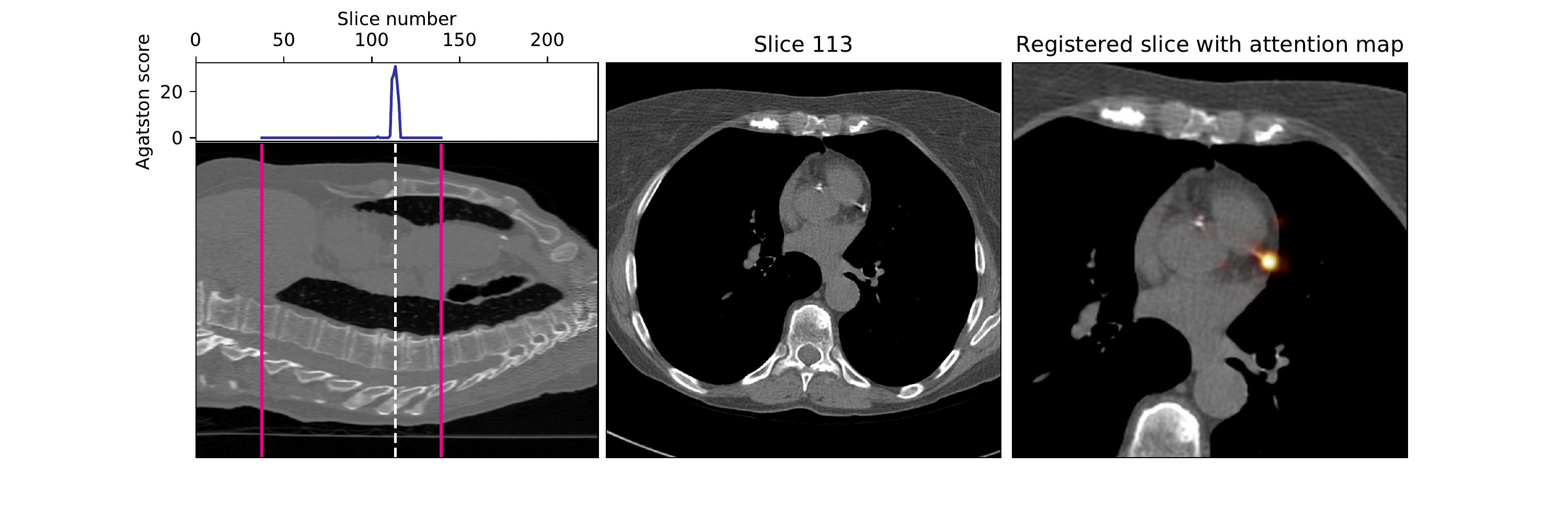}}\\
    \subfloat[Predicted: 383 -- Reference: 385]{\label{fig:attention:cardiac2}\includegraphics[height=3.2cm,trim={2.2cm 1cm 3cm 0cm},clip]{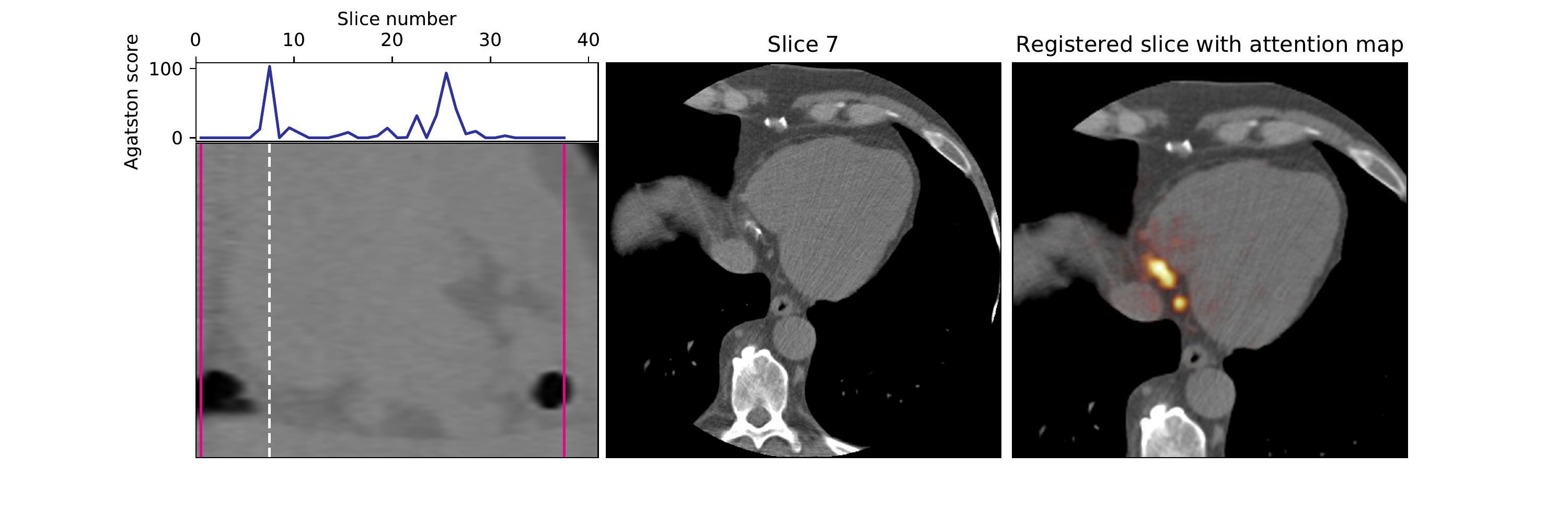}}
    \subfloat[Predicted: 229 -- Reference: 230]{\label{fig:attention:chest2}\includegraphics[height=3.2cm,trim={2.2cm 1cm 3cm 0cm},clip]{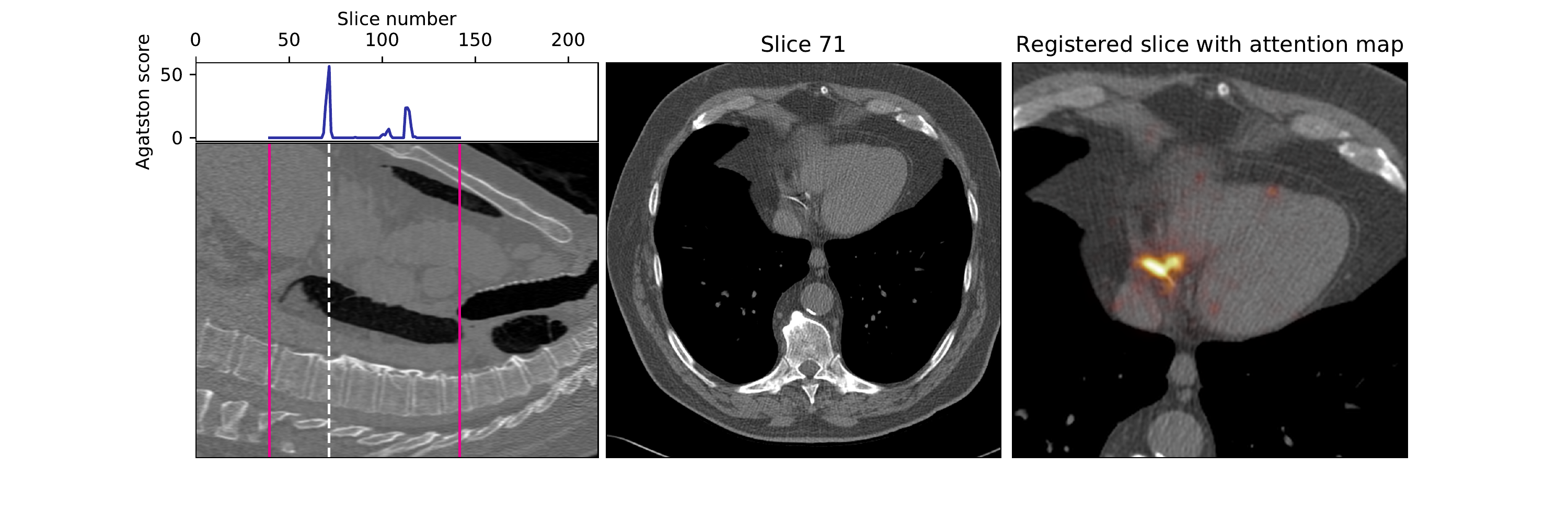}}\\
    \subfloat[Predicted: 1,021 -- Reference: 1,013]{\label{fig:attention:cardiac3}\includegraphics[height=3.2cm,trim={2.2cm 1cm 3cm 0cm},clip]{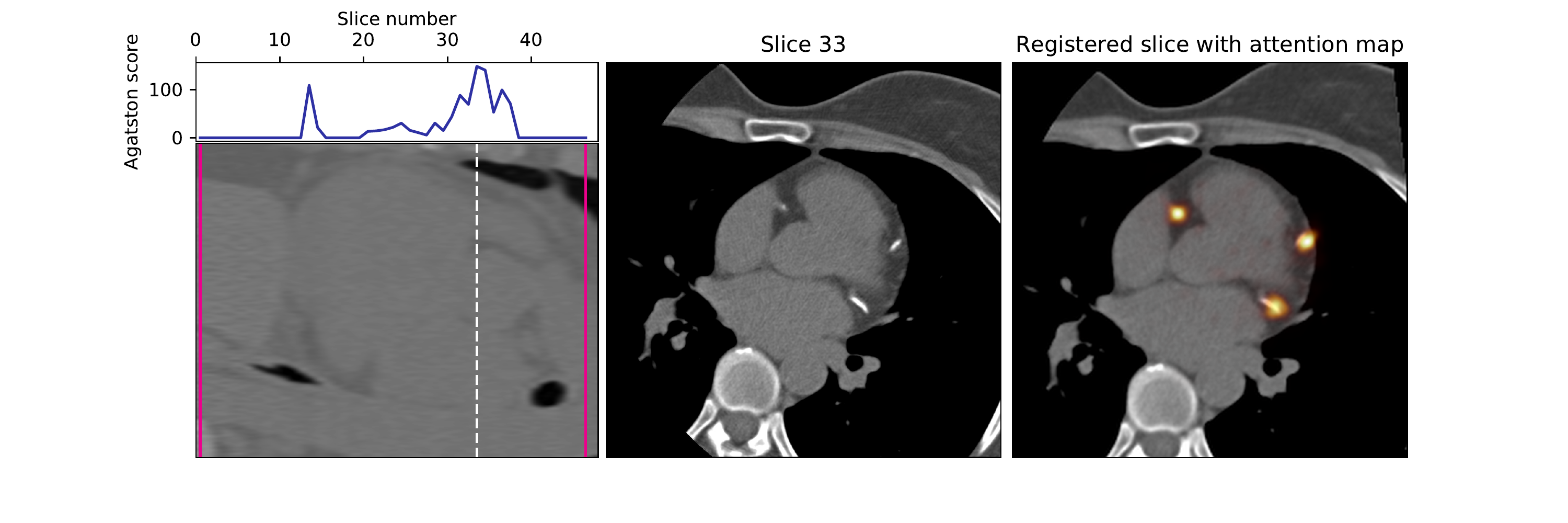}}
    \subfloat[Predicted: 436 -- Reference: 437]{\label{fig:attention:chest3}\includegraphics[height=3.2cm,trim={2.2cm 1cm 3cm 0cm},clip]{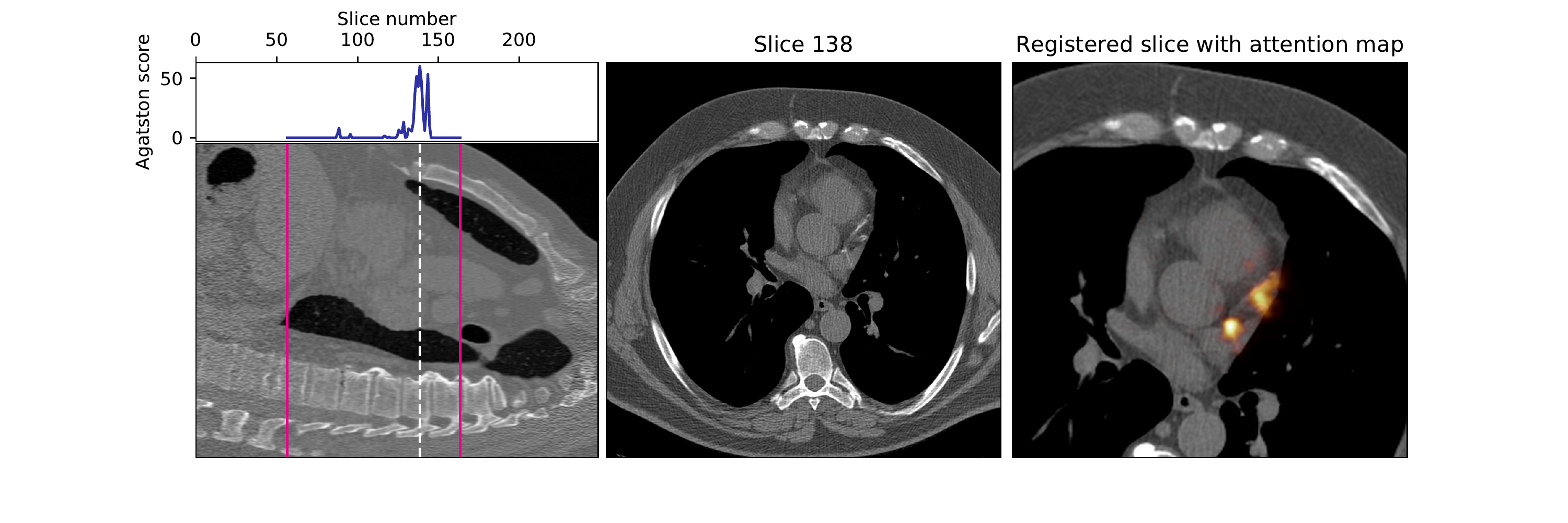}}\\
    \caption{Examples of application of decision feedback in an application of cardiac CT (left) and chest CT (right). In each example, the center sagittal slice is shown on the left with the predicted per-slice Agatston scores plotted above it. Image slices selected for further evaluation by the registration ConvNet are indicated by the solid red lines. The axial slice having the highest Agatston score is indicated with the dashed white line and is shown in the middle. The right image shows the registered slice with the resulting decision feedback superpositioned as a heatmap. In both cardiac and chest CT, decision feedback shows that the method correctly focuses on large and small calcifications in the left coronary arteries, as shown in \protect\subref{fig:attention:cardiac1} and \protect\subref{fig:attention:chest1}. Note that it is not fooled by other calcifications. Also, right coronary arteries are correctly identified, as shown in \protect\subref{fig:attention:cardiac2} and \protect\subref{fig:attention:chest2}. Even scans having extensive calcifications the method focuses correctly different locations of CAC as is shown in \protect\subref{fig:attention:cardiac3} and \protect\subref{fig:attention:chest3}.}
    \label{fig:feedback}
\end{figure*}

Decision feedback visualizes attention of the calcium scoring ConvNet. This feedback informs and end-user about the regions that contributed to the calcium score. Figure~\ref{fig:feedback} shows examples of such feedback. The feedback helps an expert to quickly navigate and evaluate the image slices containing CAC.

\begin{figure}[]
    \centering
    \includegraphics[width=.8\linewidth]{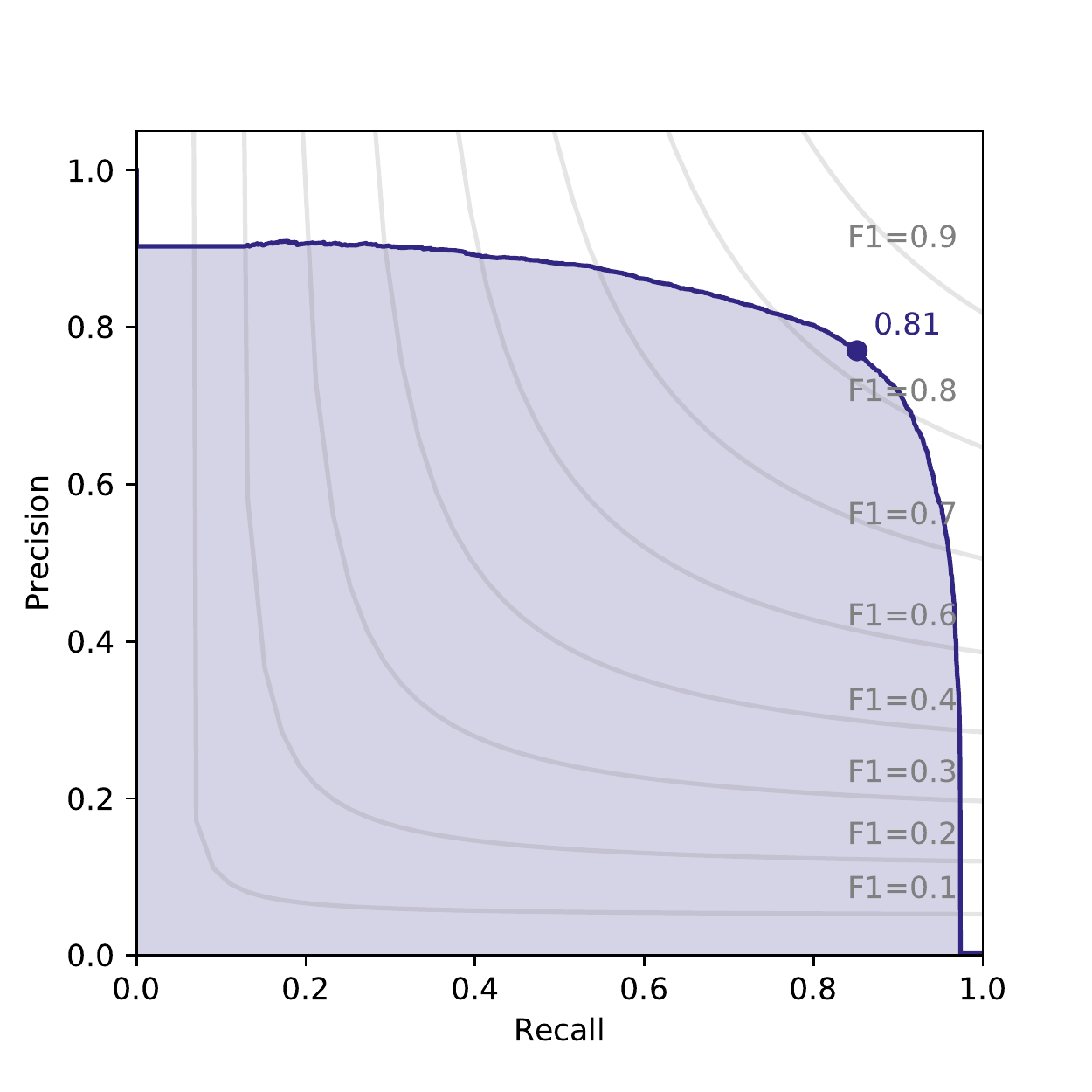}
    \caption{Precision recall curve of CAC segmentation using the obtained visual feedback heatmaps. The analysis is performed on the validation set to obtain an optimal threshold for evaluation. Optimal F1 score was 0.81 at a threshold of 0.27. Final results for quantitative evaluation of visualization feedback are shown in Table~\ref{tab:feedback_evaluation}.}
    \label{fig:precisionrecall}
\end{figure}

We propose visual feedback as an optional qualitative tool, but we have performed a quantitative analysis to provide insight in its accuracy. To obtain quantitative results we analyzed heatmaps for slices with predicted calcium scores. The heatmaps were warped to the original image spaces by using the inverse transformation matrices. The values of the heatmaps were scaled between 0 and 1 to mimic probability maps for CAC candidate voxels. CAC candidates were defined as high density 26-connected voxels with a volume between 1.5 and 1,500\,mm\textsuperscript{3}\cite{wolterink2015}. For evaluation of these maps we performed precision-recall analysis (Figure~\ref{fig:precisionrecall}). We have defined an optimal threshold by selecting the maximum F1 (i.e. Dice) score on the validation set. Table~\ref{tab:feedback_evaluation} shows the obtained scores using the selected threshold on the test sets. The results show that detection performance is very accurate on the validation set as well as the test set.

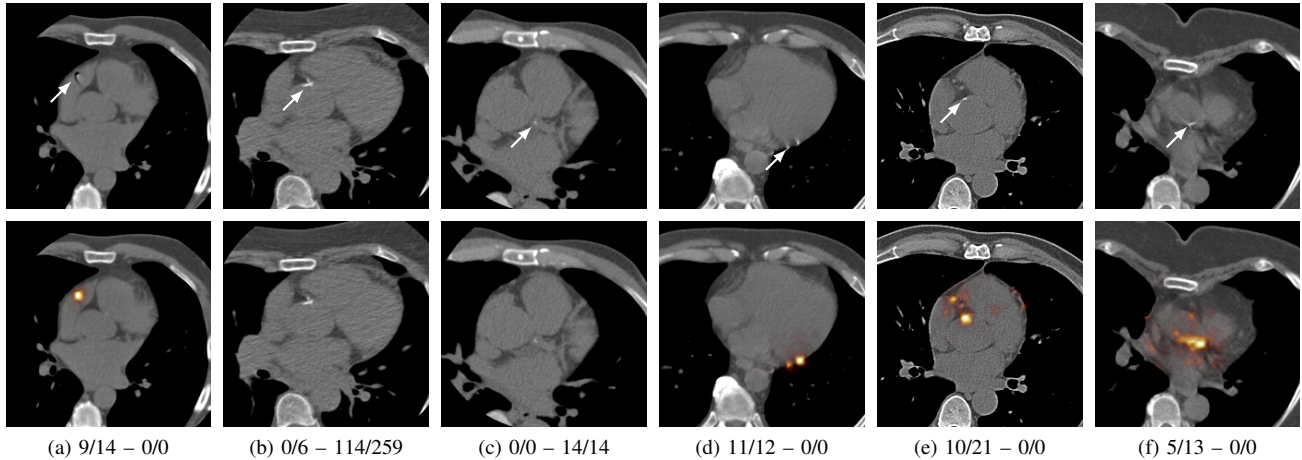
\begin{figure*}
    \centering
    \subfloat[9/14 -- 0/0]{%
        \begin{tikzpicture}
        \begin{scope}[]
            \node[anchor=south west, inner sep=0] (image) at (0,0)
             {\includegraphics[width=0.15\linewidth]{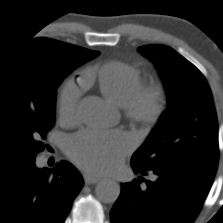}};
            \begin{scope}[x={(image.south east)}, y={(image.north west)}]
                \draw[-latex, white, thick] (0.22,0.52) to (0.32,0.62);
            \end{scope}
            \node[anchor=south west, inner sep=0] (image) at (0,-2.9) {\includegraphics[width=0.15\linewidth]{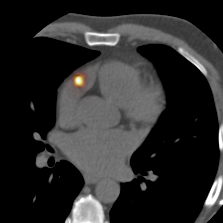}};
        \end{scope}
        
        \end{tikzpicture}%
        
        \label{fig:incorrect:cardiac1}}\enskip
    \subfloat[0/6 -- 114/259]{%
        \begin{tikzpicture}
        \begin{scope}[]
            \node[anchor=south west, inner sep=0] (image) at (0,0) {\includegraphics[width=0.15\linewidth]{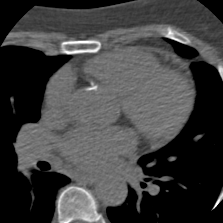}};
            \begin{scope}[x={(image.south east)}, y={(image.north west)}]
                \draw[-latex, white, thick] (0.29,0.48) to (0.39,0.58);
            \end{scope}
            \node[anchor=south west, inner sep=0] (image) at (0,-2.9) {\includegraphics[width=0.15\linewidth]{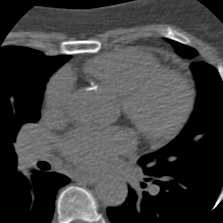}};
        \end{scope}
        \end{tikzpicture}%
        \label{fig:incorrect:cardiac2}}\enskip
    \subfloat[0/0 -- 14/14]{%
        \begin{tikzpicture}
        \begin{scope}[]
            \node[anchor=south west, inner sep=0] (image) at (0,0) {\includegraphics[width=0.15\linewidth]{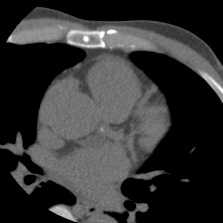}};
            \begin{scope}[x={(image.south east)}, y={(image.north west)}]
                \draw[-latex, white, thick] (0.34,0.30) to (0.44,0.40);
            \end{scope}
            \node[anchor=south west, inner sep=0] (image) at (0,-2.9) {\includegraphics[width=0.15\linewidth]{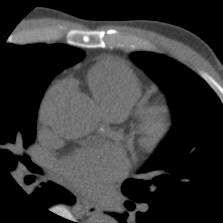}};
        \end{scope}
        \end{tikzpicture}%
        \label{fig:incorrect:cardiac3}}\enskip
    \subfloat[11/12 -- 0/0]{%
        \begin{tikzpicture}
        \begin{scope}[]
            \node[anchor=south west, inner sep=0] (image) at (0,0) {\includegraphics[width=0.15\linewidth]{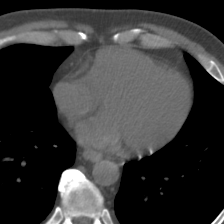}};
            \begin{scope}[x={(image.south east)}, y={(image.north west)}]
                \draw[-latex, white, thick] (0.52, 0.19) to (0.62,0.29);
            \end{scope}
            \node[anchor=south west, inner sep=0] (image) at (0,-2.9) {\includegraphics[width=0.15\linewidth]{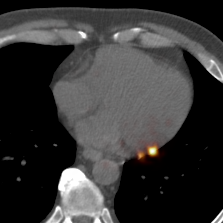}};
        \end{scope}
        \end{tikzpicture}%
        \label{fig:incorrect:chest1}}\enskip
    \subfloat[10/21 -- 0/0]{%
        \begin{tikzpicture}
        \begin{scope}[]
            \node[anchor=south west, inner sep=0] (image) at (0,0) {\includegraphics[width=0.15\linewidth]{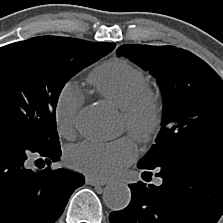}};
            \begin{scope}[x={(image.south east)}, y={(image.north west)}]
                \draw[-latex, white, thick] (0.31,0.42) to (0.41,0.52);
            \end{scope}
            \node[anchor=south west, inner sep=0] (image) at (0,-2.9) {\includegraphics[width=0.15\linewidth]{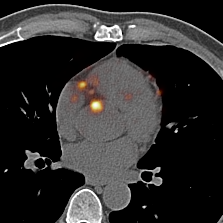}};
        \end{scope}
        \end{tikzpicture}%
        \label{fig:incorrect:chest2}}\enskip
    \subfloat[5/13 -- 0/0]{%
        \begin{tikzpicture}
        \begin{scope}[]
            \node[anchor=south west, inner sep=0] (image) at (0,0) {\includegraphics[width=0.15\linewidth]{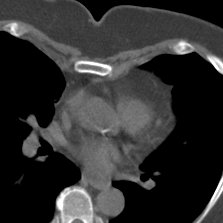}};
            \begin{scope}[x={(image.south east)}, y={(image.north west)}]
                \draw[-latex, white, thick] (0.35,0.30) to (0.45,0.40);
            \end{scope}
            \node[anchor=south west, inner sep=0] (image) at (0,-2.9) {\includegraphics[width=0.15\linewidth]{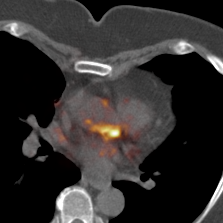}};
        \end{scope}
        \end{tikzpicture}%
        \label{fig:incorrect:chest3}}
    \caption{Examples of the largest errors, in terms of CVD risk categorization, made in cardiac CT (a-c) and in chest CT (d-f). Each image shows the axial slice most illustrative for the error. For image slices with a predicted calcium score, the heatmap is also provided. The captions show \textit{predicted slice calcium score} / \textit{predicted total calcium score} -- \textit{reference slice calcium score} / \textit{reference total calcium score}. In \protect\subref{fig:incorrect:cardiac1} a pacemaker lead, affected by a motion artifact, was incorrectly quantified as CAC. CAC near the coronary ostia was not quantified in \protect\subref{fig:incorrect:cardiac2}--having an incorrect reference annotation--and in \protect\subref{fig:incorrect:cardiac3}. In \protect\subref{fig:incorrect:chest1} infrequently occuring calcification of the pericardium was quantified as CAC. In \protect\subref{fig:incorrect:chest2} and \protect\subref{fig:incorrect:chest3} calcifications near the coronary ostia were incorrectly quantified as CAC.}
    \label{fig:incorrect}
\end{figure*}

Additionally, decision feedback aided our analysis by clarifying incorrect calcium scores. Decision feedback revealed that the largest CVD miscategorizations were not caused by incorrect quantification but by incorrect recognition of CAC. Figure~\ref{fig:incorrect} shows six examples of the largest miscategorizations made by the calcium scoring ConvNet. The majority of errors were made in identification of calcifications near the coronary artery ostia. Calcifications near the ostia can be partly in the aorta and partly in the coronary artery. These calcifications are difficult to distinguish, especially when no information of neighboring slices is available. 

\begin{table}[]
    \centering
    \caption{Quantitative evaluation of visual feedback. Evaluation was performed segmenting CAC lesions with the visualization feedback. An optimal threshold was selected using precision recall analysis on the validation data shown in Figure~\ref{fig:precisionrecall}. Final results show that visualization by the heatmap is is as accurate on the validation as on the test set.}
    \label{tab:feedback_evaluation}
    \begin{tabular}{l|cc}
        & Cardiac CT & Chest CT \\
        \hline
        Precision & 0.77 & 0.78\\
        Recall & 0.85 & 0.86\\
        Accuracy & 0.99 & 0.99\\
        F1 (Dice) score  & 0.81 & 0.82\\
    \end{tabular}
\end{table}

\subsection{Influence of training data and registration}
\begin{figure}
    \color{black}
    \centering
    \includegraphics[width=\columnwidth]{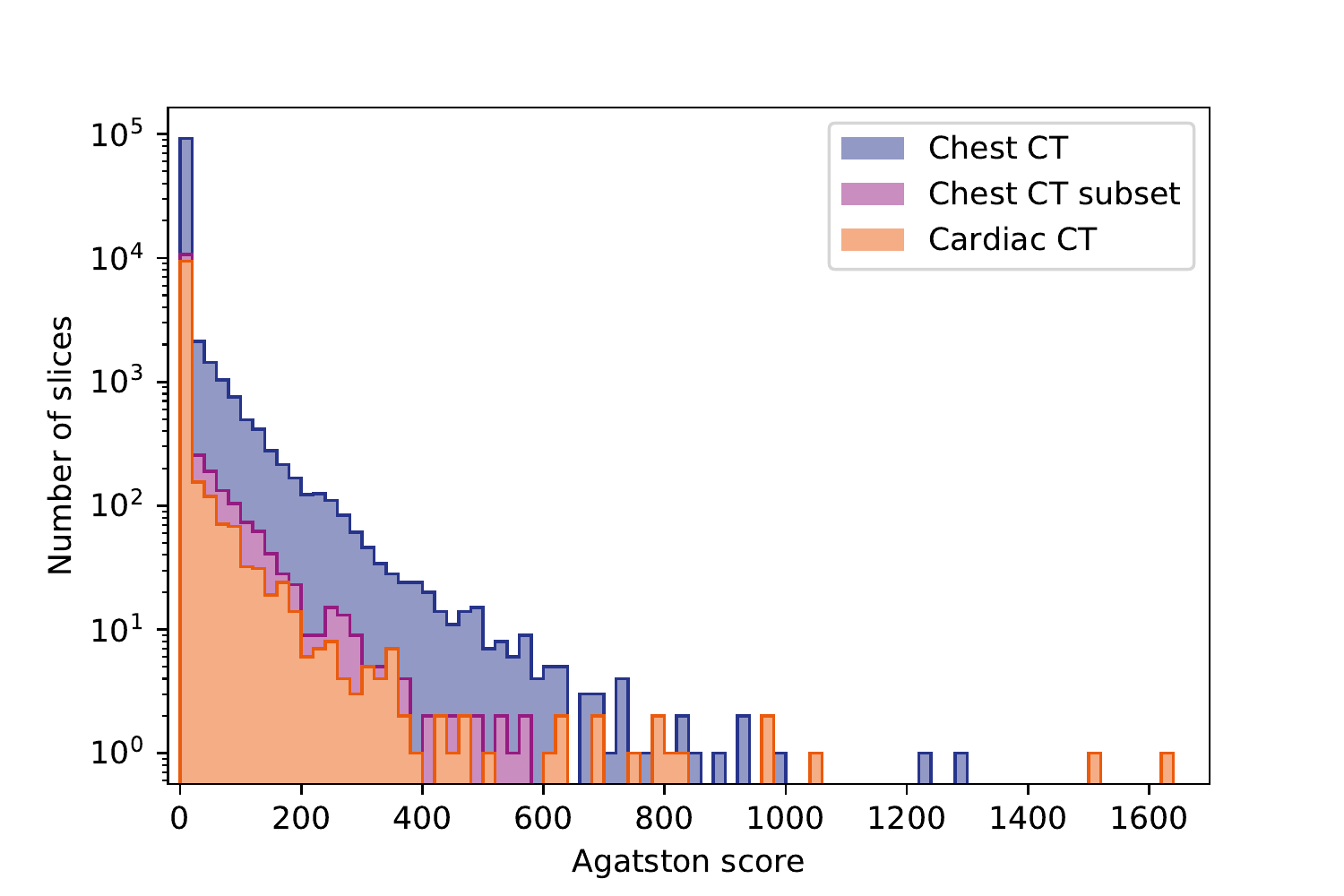}
    \caption{Histograms of per slice Agatston scores of the registered training datasets. Note that Agatston scores shown here are not corrected by factor~$\frac{i_s}{t_s}$. Please see Section~\ref{sec:cacscoremethod} for application of this correction factor in the Agatston score.}
    \label{fig:histograms}
\end{figure}

For clinical application it would be useful to investigate whether the method needs training data from both datasets or if data from one set would suffice, and we investigated the influence of atlas-registration is required. Thus, we performed experiments using different combinations of training data with and without atlas-registration, as listed in Table~\ref{tab:allexperiments}. 
The calcium scoring ConvNets were trained with either cardiac CT images, chest CT images, or a combination thereof. To balance cardiac and chest CT data, a subset of chest CT images was created by taking images from 237 randomly selected subjects and by removing every other slice in the chest CT images. Additionally, the histograms shown in Figure~\ref{fig:histograms} provide insight in the distribution of calcium amount in the training data. Note that the chest CT subset has a very similar distribution compared to the cardiac CT training set.

The best performance was achieved using atlas-registration with a calcium scoring ConvNet trained on all cardiac and chest CT images.
Lower scores are found when a calcium scoring ConvNet is only trained with cardiac CT or the subset of chest CTs. However, combining the two datasets increased the scores notably, giving a performance close to the ConvNet trained with all images. Furthermore, the results show that atlas-registration facilitated training on one type of data and high performance on the other: the ConvNet trained with the full set of chest CTs achieved a high performance on the cardiac CT test images that was very close to the best results.

\begin{table*}
    \centering
    \caption{Comparison of direct calcium scoring experiments using various datasets with and without atlas-registration. The number of CTs in the training set as well as the number of slices are given for each experiment. Additionally we provide the fraction of slices having a calcium score $>0$. The results indicate that a calcium scoring ConvNet can be trained on one type of data and evaluated on another.}
    \label{tab:allexperiments}
    \begin{tabular}{cccccc|ccc|ccc}
    \multicolumn{6}{c}{}&\multicolumn{6}{c}{\textit{Evaluated on:}}\\
        \multicolumn{6}{c}{}&\multicolumn{3}{c}{Cardiac CT}&\multicolumn{3}{c}{Chest CT}\\
        & & Data & CTs & Slices & Fraction CAC & $\kappa$ & Acc. & ICC & $\kappa$ & Acc. & ICC \\ 
        \cline{2-12}
        \multirow{8}{*}{\rotatebox[origin=c]{90}{\textit{Trained on:}}}
        &\multirow{3}{*}{Non-Registered} & Cardiac CT    &   237 &10,468 & 10.4\%  & 0.92 & 0.89 & 0.89                       & 0.46 & 0.41 & 0.24 \\
        && Chest CT                  & 1,012 &211,353 & \phantom{0}6.6\% & 0.48 & 0.59 & 0.24                       & 0.91 & 0.86 & 0.93 \\
        && Cardiac + Chest CT        & 1,239 &221,821 & \phantom{0}6.7\%& 0.90 & 0.86  & 0.87                       & 0.92 & 0.88 & 0.94 \\
        
        \cline{2-12}
        &\multirow{5}{*}{Registered}& Cardiac CT                &   237 &10,016   & 10.9\% & 0.92 & 0.88 & 0.97                       & 0.86 & 0.79 & 0.90 \\
        && Chest CT subset           &   237 &11,716   & 14.8\%& 0.91 & 0.86 & 0.95                       & 0.90 & 0.85 & 0.93 \\
        && Cardiac + Chest CT subset &   574 &21,732   & 13.0\%& 0.94 & 0.92 & \textbf{0.99}     & 0.91 & 0.88 & 0.97 \\
        && Chest CT                  & 1,012 &100,379  & 13.8\% & 0.94 & 0.91 & 0.98                       & \textbf{0.93} & 0.89 & \textbf{0.98} \\
        && Cardiac + Chest CT        & 1,239 &110,395 & 13.5\% & \textbf{0.95} & \textbf{0.93} & 0.98     & \textbf{0.93} & \textbf{0.90} & \textbf{0.98} \\
    \end{tabular}
\end{table*}

\subsection{Comparison with other methods}
Table~\ref{tab:comparison} shows a comparison with other state-of-the-art calcium scoring methods by Wolterink et al.~\cite{wolterink2015} and Lessmann et al.~\cite{lessmann2018} using the same datasets. The proposed method achieves similar performance compared to these methods, but it is hundreds of times faster. Even when ran on a single core of a CPU, the method achieves high speed. Additionally, we listed results from other direct calcium scoring methods by Gonz\'{a}lez et al. \cite{gonzalez2016} and Cano-Espinosa et al. \cite{cano2018} using chest CT data from the COPDGene study~\cite{regan2010}. We provide similar performance metrics to give an indication, but please note that a direct comparison between these methods and ours was not possible.

\begin{table*}
\centering
\caption{Results of state-of-the-art automatic calcium scoring methods in cardiac CT and chest CT, direct calcium scoring methods in chest CT, and our proposed method. For each of the studies the number of scans are given that were used for evaluation. To allow better comparison, similar statistics are reported as described in the other studies: ICC quantifies agreement between automatic scores and manual reference scores, and correlation is reported with Pearson's $\rho$. Additionally, linearly weighted $\kappa$ and accuracy are reported for three different stratifications into risk categories: (A) five categories as used in \cite{isgum2012} and \cite{wolterink2015} \{$<1$, $[1, 10)$, $[10, 100)$, $[100, 400)$, $\geq400$\}; (B) four categories as used in~\cite{lessmann2018} \{$\leq10$, $(10, 100]$, $(100, 1000]$, $>1000$\}; and (C) five categories as used in~\cite{gonzalez2016} and \cite{cano2018} \{$[0, 10)$, $[10, 100)$, $[100, 400)$, $[400, 1000)$ $\geq1000$\}. Methods evaluated on identical datasets can be compared directly. The methods by Wolterink et al. \cite{wolterink2015}, Lessmann et al. \cite{lessmann2018}, and our proposed method were evaluated on systems with an Intel Xeon E5-1620 CPU, 32 GB of internal memory, and an NVIDIA Titan X GPU.}
\label{tab:comparison}
    \begin{tabular}{llc|cc|cc|cc|cc|cc}
                                  &            \multicolumn{2}{c|}{Data}    &     \multicolumn{2}{c|}{Correlation}          & \multicolumn{2}{c|}{A}          & \multicolumn{2}{c|}{B}          & \multicolumn{2}{c|}{C}           &        \multicolumn{2}{c}{Execution time}      \\
                                                & Source & Number & ICC  & $\rho$ & $\kappa$  & acc. & $\kappa$  & acc. & $\kappa$ & acc. & CPU & GPU\\
                                                
        \hline
        \hline
                        
        \multicolumn{13}{l}{Cardiac CT}\\
        \hline
        Wolterink et al.\cite{wolterink2015} & UMCU        & 530      & 0.96 &   --    & 0.95   & 0.91     &  --  &    --  &    --  &   --       & 20\,min  &     -- \\
    Proposed method                          & UMCU        & 530      & 0.97 & 0.99    & 0.95   & 0.93     & 0.95   & 0.96     & 0.94  & 0.93     & 5\,s     & 0.15\,s  \\
    \hline
    \hline
    \multicolumn{13}{l}{Chest CT}\\
    \hline
    % Gonz\'{a}lez et al.~\cite{gonzalez2016}  &  COPDGene        & 1,000   &   --  & 0.86      &   --    &  --       & --      &  --    & 0.72  & 0.68     &  --       & 2\,min   \\
    Cano-Espinosa et al.~\cite{cano2018}     & COPDGene         & 1,000   &   --  & 0.93      &   --    &  --       & --      &  --    & 0.80  & 0.76     &   --   &  --       \\
    Lessmann et al.~\cite{lessmann2018}      & NLST             & 506     &   --  &     --    &   --    &  --       & 0.91   & 0.91     & --     &   --   & --        & 7\,min   \\
    Proposed method                          & NLST             & 506     & 0.98   & 0.97     & 0.93   & 0.90     & 0.92   & 0.91     & 0.93  & 0.90     & 11\,s    & 0.30\,s \\
\end{tabular}
\end{table*}
\subsection{Performance on orCaScore data}
We evaluated our method on data from the orCaScore challenge~\cite{wolterink2016orcascore}. This challenge provides data to evaluate a method for coronary calcium scoring. The data consists of non-contrast enhanced ECG-triggered cardiac CT acquired on CT scanners from four different vendors from four different hospitals. Training data is provided, but we evaluated our method on the test set of 40 patients without retraining. Table~\ref{tab:orcascoreresult} shows the obtained confusion matrix and lists the results of dedicated cardiac CT calcium scoring methods that competed in the challenge. Given that our method does not differentiate between location of CAC, we only provide total calcium scoring results.

\begingroup
\begin{table}
    \centering
    \caption{Results of the proposed method on orCaScore challenge data. Left: The confusion matrix shows agreement in CVD risk categorization based on the total Agatston scores: \rom{1}: $0$, \rom{2}: $[1, 100)$, \rom{3}: $[100, 300)$, \rom{4}:$>300$. The corresponding linearly weighted $\kappa$ is shown below the confusion matrix. Right: Comparison with other methods evaluated in the challenge~\cite{wolterink2016orcascore}.}
    \label{tab:orcascoreresult}
    \setlength\tabcolsep{4pt}
    \begin{tabular}{rr|cccc}
    \multicolumn{2}{c}{}&\multicolumn{4}{c}{\textit{Predicted}}\\
    {} &{} &    \rom{1} &  \rom{2} &  \rom{3} &  \rom{4} \\
    \cline{2-6}
    \multirow{5}{*}{\rotatebox[origin=c]{90}{\textit{Reference}}}
    &\rom{1} &      \textbf{8} &       0 &        0 &        0\\
    &\rom{2} &       0 &       \textbf{12} &       0&        0\\
    &\rom{3} &        0 &       0&       \textbf{8} &        0\\
    &\rom{4} &        0 &        0 &       1 &       \textbf{11}\\
    \multicolumn{2}{c}{}&\multicolumn{4}{c}{$\kappa=0.98$}
    \end{tabular}\qquad
    \begin{tabular}{c|ccc}
        \multicolumn{4}{c}{}\\
         Method & $\kappa$ & Acc. & ICC \\ 
         \hline
          A\cite{shahzad2013} &0.88&0.85&0.97\\
          B\cite{wolterink2016orcascore} &0.98&0.98&0.99\\
          C\cite{wolterink2016orcascore} &0.96&0.95&0.98\\
          D\cite{wolterink2016orcascore} &0.80&0.80&0.60\\
          E\cite{wolterink2015} &1.00&1.00&0.99\\
        %  \hline
         Ours                & 0.98 & 0.98 & 0.98\\
    \end{tabular}
\end{table}
\endgroup

{\color{black}%
\subsection{Per-artery calcium scores}
Routine coronary artery calcium scoring is typically performed per artery. Currently, only total coronary calcium scores are reported and used for CVD risk prediction. For research purposes, per-artery calcium scores might provide interesting additional information. Hence, we evaluated performance of the proposed method for per-artery calcium scoring, i.e. scoring in the the LAD, LCX, and RCA. We chose to combine CAC scores in the LM and LAD, since it is difficult, if not impossible, to differentiate them in chest CT scans. The direct scoring ConvNet was adapted by changing the number of output nodes from one to to three. Similar to the experiment described in Section~\ref{sec:calciumscoringconvnet}, training started with a balanced set of image slices with and without calcium scores for the first 10,000 iterations and continued with the full set of image slices thereafter. Additionally, each mini-batch had at least three image slices containing each type of arterial calcification. Risk categories are clinically not defined for per-artery calcium scores, but they are obtained for total calcium scores by summation of per-artery scores. The results are listed in Table~\ref{tab:perarteryscores}.

\begin{table}
\color{black}
\centering
\caption{Intraclass correlation coefficient (ICC) for per-artery calcium scores. Since CVD risk categories are not defined for per-artery scores, CVD risk categorization was evaluated with linearly weighted $\kappa$ and accuracy (Acc.) on the total calcium scores obtained by summation.}
\label{tab:perarteryscores}
\begin{tabular}{l|ccc|ccc}
        & \multicolumn{3}{c|}{Per-artery ICC} & \multicolumn{3}{c}{Total scores} \\
        & LAD & LCX  & RCA  & $\kappa$ & Acc. & ICC \\
        \hline
Cardiac CT & 0.93   & 0.88 & 0.97 & 0.94 & 0.91 & 0.97 \\
Chest CT  & 0.91   & 0.80 & 0.98 & 0.92 & 0.88 & 0.96 \\
\end{tabular}
\end{table}
}

\section{Discussion}
We have presented a method for automatic coronary calcium scoring in cardiac CT and chest CT. The method uses an atlas-registration ConvNet to align FOVs making input images alike. The atlas-registration ConvNet is trained for 3-D registration, but its rigid model is constrained to enable 2-D slice selection and 2-D image warping. Selected and warped input image slices are presented to a calcium scoring ConvNet that directly predicts the Agatston score in these slices. The method circumvents time-consuming CAC segmentation. To provide decision feedback, a visual attention heatmap can be generated that shows the regions in an image contributing to the calcium score.
The method achieves excellent agreement for calcium score prediction for CVD risk categorization compared to manual calcium scoring. The method achieves similar performance compared to state-of-the-art methods, but achieves it hundreds of times faster.

In preliminary experiments we found that only a small ConvNet architecture was able to learn direct calcium scoring. Large ConvNet architectures architectures were unstable and failed to converge during training. By limiting the degrees of freedom of a ConvNet, i.e. by using a small architecture, we were able to train a ConvNet that learned to differentiate coronary calcification from other types of 
calcification e.g. aorta calcification, pericardium calcification, and heart valve calcification.

To simplify the problem we extracted bounding boxes around the heart in our preliminary work~\cite{devos2017rsna, devos2017arxiv}. However, this was a supervised method that classified presence of the heart in image slices. In case of noisy images, consecutive image slices could have discontinuous predictions. Discontinuous predictions resulted in an incorrect bounding box extracting a partial heart. For atlas-registration used in our current work this is not an issue.

The atlas-registration ConvNets were highly successful in pre-alignment of input CTs, i.e. in slice selection and image warping. Only 4 out of 1,036 test images had slices containing CAC that were missed by erroneous slice selection. Erroneous slice selection was likely caused by incorrect focus of the atlas-registration ConvNet on high contrast areas like the diaphragm. A mask drawn around the heart might steer focus of the ConvNet and might increase registration performance. Alternatively, a simple adjustment could be made by padding slice selection with some slices. Nonetheless, the errors caused by registration had negligible impact on calcium scoring and did not affect CVD risk categorization. Calcium scoring is better with atlas-registration than without it. Moreover, registration allows training and application of direct calcium scoring on datasets with different FOVs.

In general accuracy of predicted Agatston scores was high. Although Bland-Altman analysis showed that the method underestimated subjects with high Agatston scores. In fact, this was by design, because the method estimates a log transformed Agatston score, which induces relatively low precision for higher scores, and high precision for lower scores. Because the clinically used CVD risk categories are based on exponentially increasing Agatston scores, it is obviously more important to differentiate between subjects at low to moderate risk, than to differentiate between subjects at high risk. Thus, we imposed this higher precision on lower Agatston scores. Still, the largest CVD miscategorizations were found in the lower risk categories. Miscategorization was predominantly caused by incorrect identification of CAC and aortic calcifications near the coronary artery ostia. Even manual classification of these calcifications can be very difficult when they spread from the aorta through an ostium into the coronary artery. It often involves inspecting multiple adjacent slices in 3-D. Thus, performance of the method might be improve by exploiting additional 3-D information in future work. Additionaly, performance might improve by increasing input image resolution. The current resolution was chosen based on the majority of chest CT images, being roughly half the resolution of cardiac CTs. Nevertheless, even though all cardiac CTs were downsampled a high performance was obtained in these CTs.

The proposed method shows near perfect agreement in CVD risk categorization compared to manual calcium scoring, even when trained with a relatively low number of scans from a single dataset. Interestingly, training on one type of data allowed the model to be applied to the other type of data without any modifications or transfer learning.
However, we found that a model trained on only chest CT led to better results than a model trained only on cardiac CT. One potential reason for this may be the distribution of CAC in the datasets: the population of ex-heavy smokers typically have more CAC~\cite{jacobs2010} than the population undergoing calcium scoring cardiac CT. However, Figure~\ref{fig:histograms} shows that the distribution of CAC in equally sized datasets of cardiac CT and chest CT is similar. An alternative reason could be the presence of motion artifacts, which are nearly absent in ECG-synchronized cardiac CT, but abundant in non-ECG-synchronized chest CT. Therefore, a model trained on chest CT may be more robust to such artifacts. 
While our experiments indicated that a ConvNet trained on the cardiac and chest CT datasets supplement each other, a calcium scoring ConvNet trained with only chest CTs almost matched performance of the best performing ConvNet. Additionally, we have shown that the method obtained near perfect CVD risk categorization results on cardiac CTs from the orCaScore challenge. The method did not require retraining on representative data from the different hospitals and vendors. Having a single system that can handle potentially any CT scan that visualizes the heart would be very practical in a routine radiology setting. In future work we will investigate whether the method could be readily applied on other types CTs, without requiring retraining or fine-tuning.

Additionally, we have shown that the method can provide per-artery calcium scores. While this is not required for CVD risk categorization, it might be interesting for clinical research. In terms of ICC~\cite{koo2016}, per-artery calcium scoring achieved \textit{good reliability} ($>0.75$) in the LCX, and \textit{excellent reliability} ($>0.90$) in the LAD and the RCA. In addition, determination of CVD risk using combined per-artery scores led to \textit{almost perfect agreement} (${\kappa>0.90}$)~\cite{mchugh2012}. Nevertheless, performance was slightly better when total calcium was directly determined. This difference in performance may be a consequence of increased complexity of the per-artery scoring task while using the same number of samples for training.

The proposed method can achieve a calcium score hundreds of times faster than previously proposed methods. This is mainly due to one-shot (i.e. non-iterative) registration, and direct quantification using regression. The direct calcium scoring method circumvents time-consuming intermediate segmentation. The method might also be suitable for e.g. determination of volume, (pseudo-)mass, or number of CAC; and for quantification of other lesions or diverse anatomical structures. However, the benefit of using a segmentation approach over direct scoring is that it provides immediate insight to the end-user.
We mitigate this shortcoming of direct scoring, by providing decision feedback with a visual attention heatmap. In this way valuable feedback is still provided whenever an end-user requires it.

\section{Conclusion}
We have presented an automatic method for direct calcium scoring in cardiac CT and chest CT. The method employs two ConvNets, one for atlas-registration to align the FOV of input images to an atlas image made from cardiac CTs and one for direct calcium scoring of input image slices using regression. The method achieves robust and accurate predictions of calcium scores in real-time. By providing visual feedback, insight is given in the decision process, making it readily implementable in a clinical and research settings.

\ifCLASSOPTIONcaptionsoff
\newpage
\fi


% Generated by IEEEtran.bst, version: 1.12 (2007/01/11)
\begin{thebibliography}{10}
\providecommand{\url}[1]{#1}
\csname url@samestyle\endcsname
\providecommand{\newblock}{\relax}
\providecommand{\bibinfo}[2]{#2}
\providecommand{\BIBentrySTDinterwordspacing}{\spaceskip=0pt\relax}
\providecommand{\BIBentryALTinterwordstretchfactor}{4}
\providecommand{\BIBentryALTinterwordspacing}{\spaceskip=\fontdimen2\font plus
\BIBentryALTinterwordstretchfactor\fontdimen3\font minus
  \fontdimen4\font\relax}
\providecommand{\BIBforeignlanguage}[2]{{%
\expandafter\ifx\csname l@#1\endcsname\relax
\typeout{** WARNING: IEEEtran.bst: No hyphenation pattern has been}%
\typeout{** loaded for the language `#1'. Using the pattern for}%
\typeout{** the default language instead.}%
\else
\language=\csname l@#1\endcsname
\fi
#2}}
\providecommand{\BIBdecl}{\relax}
\BIBdecl

\bibitem{gbd2016}
{GBD 2015 Mortality and Causes of Death Collaborators}, ``Global, regional, and
  national life expectancy, all-cause mortality, and cause-specific mortality
  for 249 causes of death, 1980-2015: a systematic analysis for the global
  burden of disease study 2015,'' \emph{Lancet}, vol. 388, no. 10053, pp.
  1459--1544, Oct 2016.

\bibitem{whofactsheet}
\BIBentryALTinterwordspacing
{World Health Organization}, ``{Cardiovascular} diseases ({CVDs}) [fact
  sheet].''
\BIBentrySTDinterwordspacing

\bibitem{yeboah2012}
J.~Yeboah, R.~McClelland, T.~Polonsky, and et~al., ``Comparison of novel risk
  markers for improvement in cardiovascular risk assessment in
  intermediate-risk individuals,'' \emph{JAMA}, vol. 308, no.~8, pp. 788--795,
  2012.

\bibitem{hecht2015}
H.~S. Hecht, ``Coronary artery calcium scanning: Past, present, and future,''
  \emph{JACC: Cardiovascular Imaging}, vol.~8, no.~5, pp. 579 -- 596, 2015.

\bibitem{hecht2017}
H.~S. Hecht, P.~Cronin, M.~J. Blaha, M.~J. Budoff, E.~A. Kazerooni, J.~Narula,
  D.~Yankelevitz, and S.~Abbara, ``2016 scct/str guidelines for coronary artery
  calcium scoring of noncontrast noncardiac chest ct scans: A report of the
  society of cardiovascular computed tomography and society of thoracic
  radiology,'' \emph{Journal of Thoracic Imaging}, vol.~32, no.~5, p.
  W54–W66, 2017.

\bibitem{einstein2010}
A.~J. Einstein, L.~L. Johnson, S.~Bokhari, J.~Son, R.~C. Thompson, T.~M.
  Bateman, S.~W. Hayes, and D.~S. Berman, ``Agreement of visual estimation of
  coronary artery calcium from low-dose ct attenuation correction scans in
  hybrid {PET}/{CT} and {SPECT}/{CT} with standard agatston score,''
  \emph{Journal of the American College of Cardiology}, vol.~56, no.~23, pp.
  1914--1921, Nov 2010.

\bibitem{mylonas2012}
I.~Mylonas, M.~Kazmi, L.~Fuller, R.~A. deKemp, Y.~Yam, L.~Chen, R.~S.
  Beanlands, and B.~J.~W. Chow, ``Measuring coronary artery calcification using
  positron emission tomography-computed tomography attenuation correction
  images,'' \emph{European Heart Journal Cardiovascular Imaging}, vol.~13,
  no.~9, pp. 786--792, Sep 2012.

\bibitem{gernaat2016}
S.~A.~M. Gernaat, I.~I\v{s}gum, B.~D. de~Vos, R.~A.~P. Takx, D.~A. Young-Afat,
  N.~Rijnberg, D.~E. Grobbee, Y.~van~der Graaf, P.~A. de~Jong, T.~Leiner, and
  et~al., ``Automatic coronary artery calcium scoring on radiotherapy planning
  {CT} scans of breast cancer patients: Reproducibility and association with
  traditional cardiovascular risk factors,'' \emph{PLOS ONE}, vol.~11, no.~12,
  p. e0167925, Dec 2016.

\bibitem{jacobs2010}
P.~C. Jacobs, M.~Prokop, Y.~van~der Graaf, M.~J. Gondrie, K.~J. Janssen, H.~J.
  de~Koning, I.~I\v{s}gum, R.~J. van Klaveren, M.~Oudkerk, B.~van Ginneken, and
  W.~P. Mali, ``Comparing coronary artery calcium and thoracic aorta calcium
  for prediction of all-cause mortality and cardiovascular events on low-dose
  non-gated computed tomography in a high-risk population of heavy smokers.''
  \emph{Atherosclerosis}, vol. 209, no.~2, pp. 455--462, 2010.

\bibitem{chiles2015}
C.~Chiles, F.~Duan, G.~W. Gladish, J.~G. Ravenel, S.~G. Baginski, B.~S. Snyder,
  S.~DeMello, S.~S. Desjardins, R.~F. Munden, and {NLST Study Team},
  ``Association of coronary artery calcification and mortality in the national
  lung screening trial: A comparison of three scoring methods,''
  \emph{Radiology}, vol. 276, no.~1, pp. 82--90, 2015.

\bibitem{nlst2011b}
{The National Lung Screening Trial Research Team}, ``Reduced lung-cancer
  mortality with low-dose computed tomographic screening,'' \emph{New England
  Journal of Medicine}, vol. 365, no.~5, pp. 395--409, 2011.

\bibitem{agatston1990}
A.~S. Agatston, W.~R. Janowitz, F.~J. Hildner, N.~R. Zusmer, M.~Viamonte, and
  R.~Detrano, ``Quantification of coronary artery calcium using ultrafast
  computed tomography,'' \emph{Journal of the American College of Cardiology},
  vol.~15, no.~4, pp. 827--832, 1990.

\bibitem{rumberger1999}
J.~A. Rumberger, B.~H. Brundage, D.~J. Rader, and G.~Kondos, ``Electron beam
  computed tomographic coronary calcium scanning: A review and guidelines for
  use in asymptomatic persons,'' \emph{Mayo Clinic Proceedings}, vol.~74,
  no.~3, pp. 243--252, Mar 1999.

\bibitem{shemesh2010}
J.~Shemesh, C.~I. Henschke, D.~Shaham, R.~Yip, A.~O. Farooqi, M.~D. Cham, D.~I.
  McCauley, M.~Chen, J.~P. Smith, D.~M. Libby, and et~al., ``Ordinal scoring of
  coronary artery calcifications on low-dose {CT} scans of the chest is
  predictive of death from cardiovascular disease,'' \emph{Radiology}, vol.
  257, no.~2, pp. 541--548, Nov 2010.

\bibitem{gonzalez2016}
G.~Gonz\'{a}lez, G.~R. Washko, and R.~S.~J. Est\'{e}par, ``Automated agatston
  score computation in a large dataset of non {ECG}-gated chest computed
  tomography,'' in \emph{2016 IEEE 13th International Symposium on Biomedical
  Imaging (ISBI)}, Apr 2016, pp. 53--57.

\bibitem{xie2017}
Y.~Xie, S.~Liu, A.~Miller, J.~A. Miller, S.~Markowitz, A.~Akhund, and A.~P.
  Reeves, ``Coronary artery calcification identification and labeling in
  low-dose chest {CT} images,'' in \emph{Proceedings of SPIE}, vol. 10134,
  2017, pp. 10\,134 -- 10\,134 -- 8.

\bibitem{isgum2012}
I.~I\v{s}gum, M.~Prokop, M.~Niemeijer, M.~A. Viergever, and B.~van Ginneken,
  ``Automatic coronary calcium scoring in low-dose chest computed tomography,''
  \emph{IEEE Transactions on Medical Imaging}, vol.~31, no.~12, pp. 2322--2334,
  2012.

\bibitem{shahzad2013}
R.~Shahzad, T.~van Walsum, M.~Schaap, A.~Rossi, S.~Klein, A.~C. Weustink, P.~J.
  de~Feyter, L.~J. van Vliet, and W.~J. Niessen, ``Vessel specific coronary
  artery calcium scoring: an automatic system,'' \emph{Academic Radiology},
  vol.~20, no.~1, pp. 1--9, Jan 2013.

\bibitem{wolterink2015}
J.~M. Wolterink, T.~Leiner, R.~A.~P. Takx, M.~A. Viergever, and I.~I\v{s}gum,
  ``Automatic coronary calcium scoring in non-contrast-enhanced {ECG}-triggered
  cardiac {CT} with ambiguity detection,'' \emph{IEEE Transactions on Medical
  Imaging}, vol.~34, no.~9, pp. 1867--1878, Sep 2015.

\bibitem{durlak2017}
F.~Durlak, M.~Wels, C.~Schwemmer, M.~S\"{u}hling, S.~Steidl, and A.~Maier,
  \emph{Growing a Random Forest with Fuzzy Spatial Features for Fully Automatic
  Artery-Specific Coronary Calcium Scoring}, ser. Lecture Notes in Computer
  Science.\hskip 1em plus 0.5em minus 0.4em\relax Springer, Cham, Sep 2017, pp.
  27--35.

\bibitem{wolterink2015miccai}
J.~M. Wolterink, T.~Leiner, M.~A. Viergever, and I.~I{\v{s}}gum, ``Automatic
  coronary calcium scoring in cardiac {CT} angiography using convolutional
  neural networks,'' in \emph{Medical Image Computing and Computer-Assisted
  Intervention -- MICCAI 2015}, N.~Navab, J.~Hornegger, W.~M. Wells, and
  A.~Frangi, Eds.\hskip 1em plus 0.5em minus 0.4em\relax Cham: Springer
  International Publishing, 2015, pp. 589--596.

\bibitem{wolterink2016}
J.~M. Wolterink, T.~Leiner, B.~D. de~Vos, R.~W. van Hamersvelt, M.~A.
  Viergever, and I.~I\v{s}gum, ``Automatic coronary artery calcium scoring in
  cardiac {CT} angiography using paired convolutional neural networks,''
  \emph{Medical Image Analysis}, vol.~34, pp. 123--136, Dec 2016.

\bibitem{lessmann2016}
N.~Lessmann, I.~I\v{s}gum, A.~A.~A. Setio, B.~D. de~Vos, F.~Ciompi, P.~A.
  de~Jong, M.~Oudkerk, W.~P. T.~M. Mali, M.~A. Viergever, and B.~van Ginneken,
  ``Deep convolutional neural networks for automatic coronary calcium scoring
  in a screening study with low-dose chest {CT},'' in \emph{Proceedings of
  SPIE}, G.~D. Tourassi and S.~G. Armato, Eds., vol. 9785, Mar 2016, p. 978511.

\bibitem{lessmann2018}
N.~Lessmann, B.~v. Ginneken, M.~Zreik, P.~A. de~Jong, B.~D. de~Vos, M.~A.
  Viergever, and I.~I\v{s}gum, ``Automatic calcium scoring in low-dose chest
  {CT} using deep neural networks with dilated convolutions,'' \emph{IEEE
  Transactions on Medical Imaging}, vol.~37, no.~2, pp. 615--625, Feb 2018.

\bibitem{wolterink2016orcascore}
J.~M. Wolterink, T.~Leiner, B.~D. de~Vos, J.-L. Coatrieux, B.~M. Kelm,
  S.~Kondo, R.~A. Salgado, R.~Shahzad, H.~Shu, M.~Snoeren, R.~A.~P. Takx, L.~J.
  van Vliet, T.~van Walsum, T.~P. Willems, G.~Yang, Y.~Zheng, M.~A. Viergever,
  and I.~Išgum, ``An evaluation of automatic coronary artery calcium scoring
  methods with cardiac ct using the orcascore framework,'' \emph{Medical
  Physics}, vol.~43, no.~5, pp. 2361--2373, 2016.

\bibitem{devos2017localization}
B.~D. de~Vos, J.~M. Wolterink, P.~A. de~Jong, T.~Leiner, M.~A. Viergever, and
  I.~I\v{s}gum, ``Convnet-based localization of anatomical structures in 3-d
  medical images,'' \emph{IEEE Transactions on Medical Imaging}, vol.~36,
  no.~7, pp. 1470--1481, July 2017.

\bibitem{hussain2017}
M.~A. Hussain, A.~Amir-Khalili, G.~Hamarneh, and R.~Abugharbieh,
  ``Segmentation-free kidney localization and volume estimation using
  aggregated orthogonal decision cnns,'' in \emph{Medical Image Computing and
  Computer-Assisted Intervention - MICCAI 2017}, M.~Descoteaux, L.~Maier-Hein,
  A.~Franz, P.~Jannin, D.~L. Collins, and S.~Duchesne, Eds.\hskip 1em plus
  0.5em minus 0.4em\relax Cham: Springer International Publishing, 2017, pp.
  612--620.

\bibitem{zhen2017}
\BIBentryALTinterwordspacing
X.~Zhen, H.~Zhang, A.~Islam, M.~Bhaduri, I.~Chan, and S.~Li, ``Direct and
  simultaneous estimation of cardiac four chamber volumes by multioutput sparse
  regression,'' \emph{Medical Image Analysis}, vol.~36, pp. 184 -- 196, 2017.
\BIBentrySTDinterwordspacing

\bibitem{xue2018}
\BIBentryALTinterwordspacing
W.~Xue, G.~Brahm, S.~Pandey, S.~Leung, and S.~Li, ``Full left ventricle
  quantification via deep multitask relationships learning,'' \emph{Medical
  Image Analysis}, vol.~43, pp. 54 -- 65, 2018.
\BIBentrySTDinterwordspacing

\bibitem{devos2017rsna}
B.~D. de~Vos, N.~Lessmann, P.~A. de~Jong, M.~A. Viergever, and I.~I\v{s}gum,
  ``Direct coronary artery calcium scoring in low-dose chest {CT} using deep
  learning analysis,'' The Radiological Society of North America's Annual
  Meeting, 2017.

\bibitem{devos2017arxiv}
B.~D. de~Vos, N.~Lessmann, P.~A. de~Jong, M.~A. Viergever, and I.~I\v{s}gum,
  ``Direct and real-time cardiovascular risk prediction,''
  \emph{arXiv:1712.02982 [cs]}, 2017.

\bibitem{isgum2017}
I.~I\v{s}gum, B.~D. de~Vos, J.~M. Wolterink, D.~Dey, D.~S. Berman, M.~Rubeaux,
  T.~Leiner, and P.~J. Slomka, ``Automatic determination of cardiovascular risk
  by {CT} attenuation correction maps in {Rb}-82 {PET}/{CT},'' \emph{Journal of
  Nuclear Cardiology}, Apr 2017.

\bibitem{devos2017registration}
B.~D. de~Vos, F.~F. Berendsen, M.~A. Viergever, M.~Staring, and I.~I\v{s}gum,
  ``End--to--end unsupervised deformable image registration with a
  convolutional neural network,'' in \emph{Deep Learning in Medical Image
  Analysis and Multimodal Learning for Clinical Decision Support: Third
  International Workshop, DLMIA 2017, and 7th International Workshop, ML-CDS
  2017, Held in Conjunction with MICCAI 2017, Qu{\'e}bec City, QC, Canada,
  September 14, Proceedings}.\hskip 1em plus 0.5em minus 0.4em\relax Cham:
  Springer International Publishing, 2017, pp. 204--212.

\bibitem{devos2018media}
\BIBentryALTinterwordspacing
B.~D. de~Vos, F.~F. Berendsen, M.~A. Viergever, H.~Sokooti, M.~Staring, and
  I.~Išgum, ``A deep learning framework for unsupervised affine and deformable
  image registration,'' \emph{Medical Image Analysis}, 2018.
\BIBentrySTDinterwordspacing

\bibitem{rutten2011}
A.~Rutten, I.~I\v{s}gum, and M.~Prokop, ``Calcium scoring with prospectively
  {ECG}-triggered {CT}: using overlapping datasets generated with {MPR}
  decreases inter-scan variability,'' \emph{European Journal of Radiology},
  vol.~80, no.~1, pp. 83--88, Oct 2011.

\bibitem{jongen2004}
C.~Jongen, J.~P.~W. Pluim, P.~J. Nederkoorn, M.~A. Viergever, and W.~J.
  Niessen, ``Construction and evaluation of an average {CT} brain image for
  inter-subject registration,'' \emph{Computers in Biology and Medicine},
  vol.~34, no.~8, pp. 647--662, Dec 2004.

\bibitem{ohnesorg2002}
B.~Ohnesorge, T.~Flohr, R.~Fischbach, A.~Kopp, A.~Knez, S.~Schr\"{o}der,
  U.~Sch\"{o}pf, A.~Crispin, E.~Klotz, M.~Reiser, and et~al., ``Reproducibility
  of coronary calcium quantification in repeat examinations with
  retrospectively {ECG}-gated multisection spiral {CT},'' \emph{European
  Radiology}, vol.~12, no.~6, pp. 1532--1540, Jun 2002.

\bibitem{ioffe2015}
S.~Ioffe and C.~Szegedy, ``Batch normalization: Accelerating deep network
  training by reducing internal covariate shift,'' in \emph{International
  Conference on Machine Learning}, Jun 2015, pp. 448--456.

\bibitem{clevert2016}
D.-A. Clevert, T.~Unterthiner, and S.~Hochreiter, ``Fast and accurate deep
  network learning by exponential linear units (elus),'' in \emph{International
  Conference on Machine Learning}, 2016, arXiv: 1511.07289.

\bibitem{zeilerfergus2014}
M.~D. Zeiler and R.~Fergus, ``Visualizing and understanding convolutional
  networks,'' in \emph{Computer Vision - European Conference on Computer Vision
  2014}, D.~Fleet, T.~Pajdla, B.~Schiele, and T.~Tuytelaars, Eds., vol.
  8689.\hskip 1em plus 0.5em minus 0.4em\relax Springer International
  Publishing, 2014, pp. 818--833.

\bibitem{theano2016}
{Theano Development Team}, ``{Theano: A {Python} framework for fast computation
  of mathematical expressions},'' \emph{arXiv e--prints}, vol. abs/1605.02688,
  2016.

\bibitem{lasagne2015}
S.~Dieleman, J.~Schl\"{u}ter, C.~Raffel, E.~Olson, S.~K. S{\o}nderby, D.~Nouri,
  D.~Maturana, M.~Thoma, E.~Battenberg, J.~Kelly, J.~D. Fauw, M.~Heilman, D.~M.
  de~Almeida, B.~McFee, H.~Weideman, G.~Tak\'{a}cs, P.~de~Rivaz, J.~Crall,
  G.~Sanders, K.~Rasul, C.~Liu, G.~French, and J.~Degrave, ``Lasagne: First
  release.'' 2015.

\bibitem{opencv}
G.~Bradski, ``{The OpenCV Library},'' \emph{Dr. Dobb's Journal of Software
  Tools}, 2000.

\bibitem{kingma2014}
D.~P. Kingma and J.~Ba, ``Adam: {A} method for stochastic optimization,'' in
  \emph{International Conference on Learning Representation}, 2015.

\bibitem{mchugh2012}
\BIBentryALTinterwordspacing
M.~L. McHugh, ``Interrater reliability: the kappa statistic,'' \emph{Biochem
  Med (Zagreb)}, vol.~22, no.~3, pp. 276--282, 2012.
\BIBentrySTDinterwordspacing

\bibitem{cano2018}
C.~Cano-Espinosa, G.~Gonz\'{a}lez, G.~R. Washko, M.~Cazorla, and R.~S.~J.
  Est\'{e}par, ``Automated agatston score computation in non-{ECG} gated {CT}
  scans using deep learning,'' in \emph{Proceedings of SPIE}, vol. 10574, 2018,
  pp. 10\,574 -- 10\,574 -- 6.

\bibitem{regan2010}
E.~A. Regan, J.~E. Hokanson, J.~R. Murphy, B.~Make, D.~A. Lynch, T.~H. Beaty,
  D.~Curran-Everett, E.~K. Silverman, and J.~D. Crapo, ``Genetic epidemiology
  of {COPD} ({COPDGene}) study design,'' \emph{COPD}, vol.~7, no.~1, pp.
  32--43, Feb 2010.

\bibitem{koo2016}
T.~K. Koo and M.~Y. Li, ``A guideline of selecting and reporting intraclass
  correlation coefficients for reliability research,'' \emph{J Chiropr Med},
  vol.~15, no.~2, pp. 155--163, 2016.

\end{thebibliography}
\end{document}